\documentclass[sigconf]{acmart}
\AtBeginDocument{%
  \providecommand\BibTeX{{%
    \normalfont B\kern-0.5em{\scshape i\kern-0.25em b}\kern-0.8em\TeX}}}

\setcopyright{acmcopyright}
\copyrightyear{2022}
\acmYear{2022}
\acmDOI{10.1145/3503161.3547942}

\acmConference[MM '22]{Proceedings of the 30th ACM International Conference on Multimedia}{October 10--14, 2022}{Lisboa, Portugal}
\acmBooktitle{Proceedings of the 30th ACM International Conference on Multimedia (MM '22), October 10--14, 2022, Lisboa, Portugal}
\acmPrice{15.00}
\acmISBN{978-1-4503-9203-7/22/10}
\acmSubmissionID{797}
\usepackage{graphicx}
\usepackage{amsmath}

\usepackage{amssymb}
\usepackage{booktabs}
\usepackage{makecell,multirow,diagbox}
\usepackage{subcaption}
\usepackage{epstopdf}

\usepackage{xcolor}
\definecolor{_color}{RGB}{178, 119, 0}

\usepackage{caption}
\usepackage{hyperref}
\usepackage[capitalize]{cleveref}
\crefname{section}{Sec.}{Secs.}
\Crefname{section}{Section}{Sections}
\Crefname{table}{Table}{Tables}
\crefname{table}{Tab.}{Tabs.}

\usepackage{balance}
\usepackage{caption}
\newcommand{\methodName}{SPTS}

\begin{document}
\title{SPTS: Single-Point Text Spotting}

\author{Dezhi Peng}
\affiliation{\institution{South China University of Technology}\country{}}
\author{Xinyu Wang}
\affiliation{\institution{Zhejiang University}\country{}}
\author{Yuliang Liu}
\authornote{Corresponding authors. Part of this work was done by Yuliang Liu at the Chinese University of Hong Kong.}
\affiliation{\institution{Huazhong University of Science and Technology}\country{}}
\author{Jiaxin Zhang}
\author{Mingxin Huang}
\affiliation{\institution{South China University of Technology}\country{}}
\author{Songxuan Lai}
\author{Jing Li}
\author{Shenggao Zhu}
\affiliation{\institution{Huawei Cloud Computing Technologies}\country{}}
\author{Dahua Lin}
\affiliation{\institution{Chinese University of Hong Kong}\country{}}
\author{Chunhua Shen}
\affiliation{\institution{Zhejiang University}\country{}}
\author{Xiang Bai}
\affiliation{\institution{Huazhong University of Science and Technology}\country{}}
\author{Lianwen Jin}
\authornotemark[1]
\affiliation{\institution{South China University of Technology}\country{}}

\def\authors{Dezhi Peng, Xinyu Wang, Yuliang Liu, et al}
\renewcommand{\shortauthors}{Dezhi Peng et al.}

\begin{abstract}
Existing scene text spotting (i.e., end-to-end text detection and recognition) methods rely on costly bounding box annotations (e.g., text-line, word-level, or character-level bounding boxes). For the first time, we demonstrate that training scene text spotting models can be achieved with an extremely low-cost annotation of a single-point for each instance. We propose an end-to-end scene text spotting method that tackles scene text spotting as a sequence prediction task. Given an image as input, we formulate the desired detection and recognition results as a sequence of discrete tokens and use an auto-regressive Transformer to predict the sequence. The proposed method is simple yet effective, which can achieve state-of-the-art results on widely used benchmarks. Most significantly, we show that the performance is not very sensitive to the positions of the point annotation, meaning that it can be much easier to be annotated or even be automatically generated than the bounding box that requires precise positions. We believe that such a pioneer attempt indicates a significant opportunity for scene text spotting applications of a much larger scale than previously possible. The code is available at \href{https://github.com/shannanyinxiang/SPTS}{https://github.com/shannanyinxiang/SPTS}. 
\end{abstract}




\maketitle


\section{Introduction}
In the last decades, it has been witnessed that modern Optical Character Recognition (OCR) algorithms are able to read textual content from pictures of complex scenes, which is an incredible development, leading to enormous interest from both academia and industry. The limitation of existing methods, and particularly their poorer performance on arbitrarily shaped scene text, have been repeatedly identified~\cite{liao2020masktext, liu2020abcnet, chng2019icdar2019}. This can be seen in the trend of worse predictions for instances with curved shapes, varied fonts, distortions, etc.

\begin{figure*}[t!]
    \centering
    \begin{subfigure}{0.18\linewidth}
        \includegraphics[width=\linewidth]{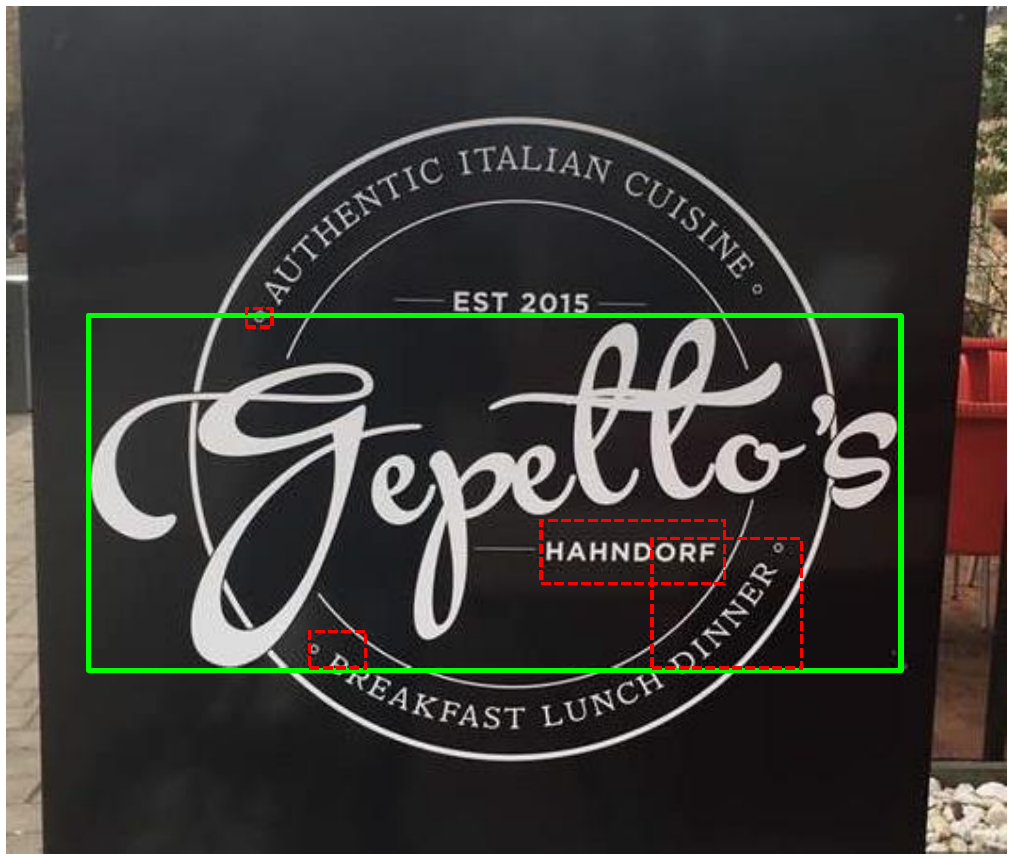}
        \caption{Rectangle (55s).} 
        \label{fig:tgt_rect}
    \end{subfigure}
    \begin{subfigure}{0.18\linewidth}
        \includegraphics[width=\linewidth]{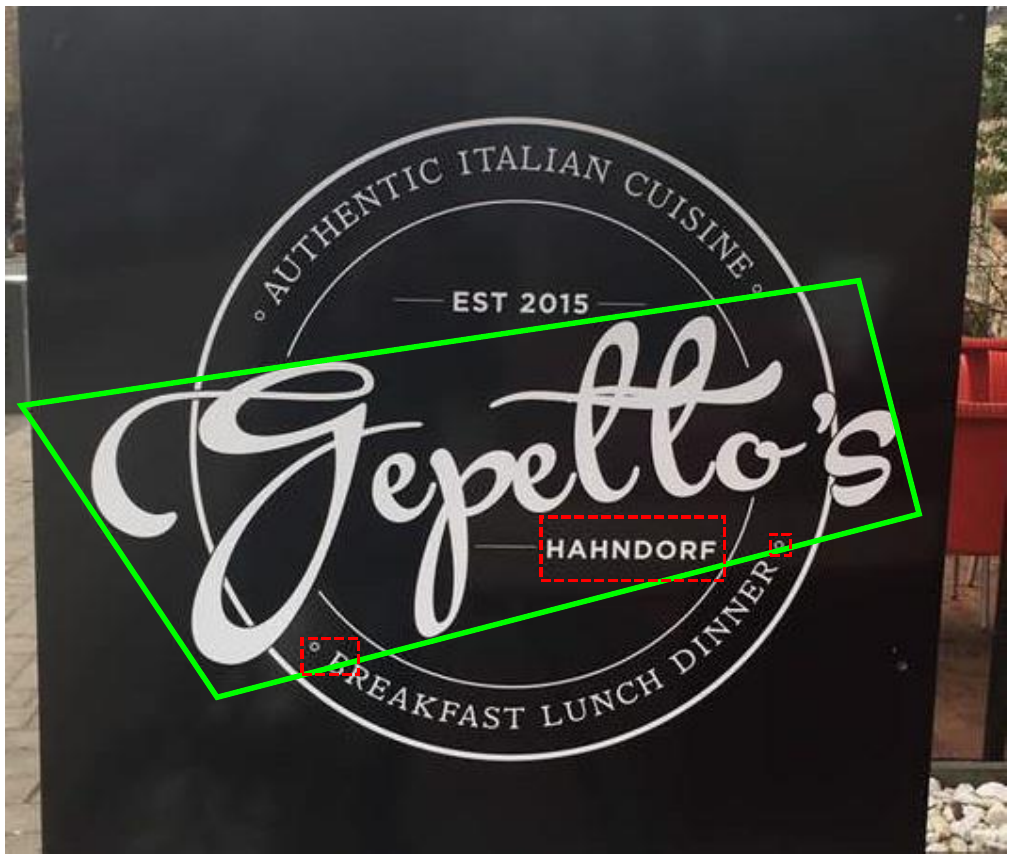}
        \caption{Quadrilateral (96s).} 
        \label{fig:tgt_quad}
    \end{subfigure}
    \begin{subfigure}{0.18\linewidth}
        \includegraphics[width=\linewidth]{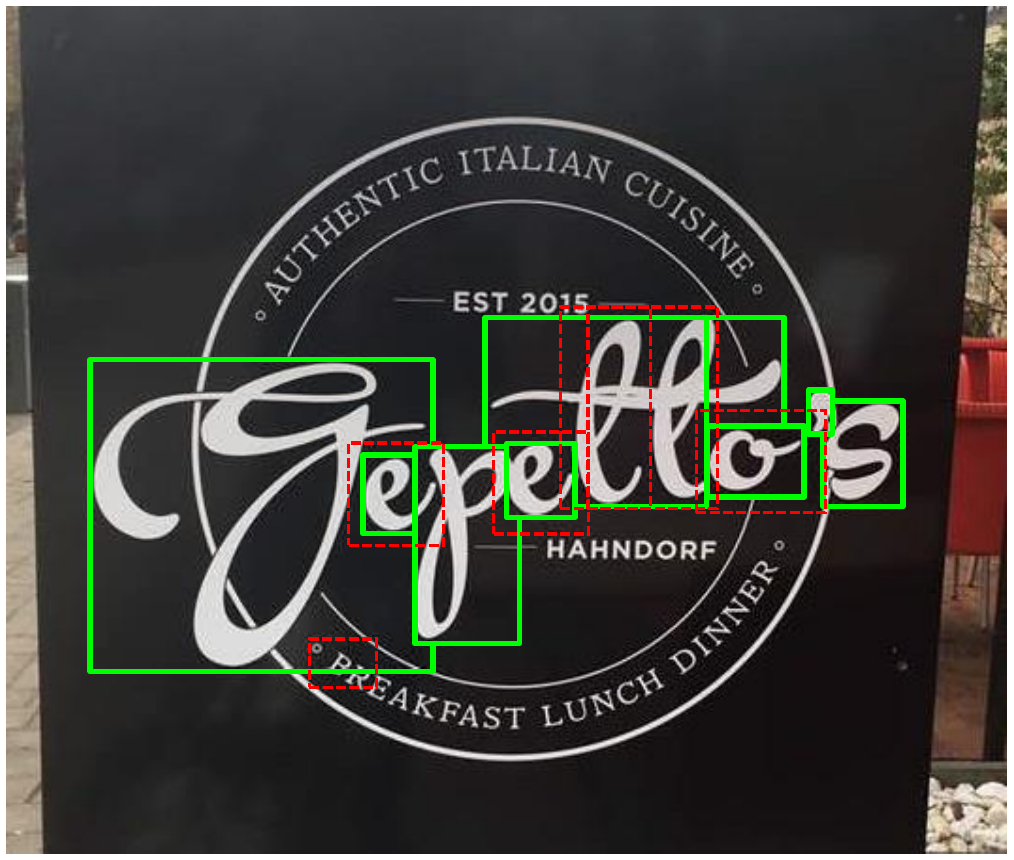}
        \caption{Character (581s).} 
        \label{fig:tgt_char}
    \end{subfigure}
    \begin{subfigure}{0.18\linewidth}
        \includegraphics[width=\linewidth]{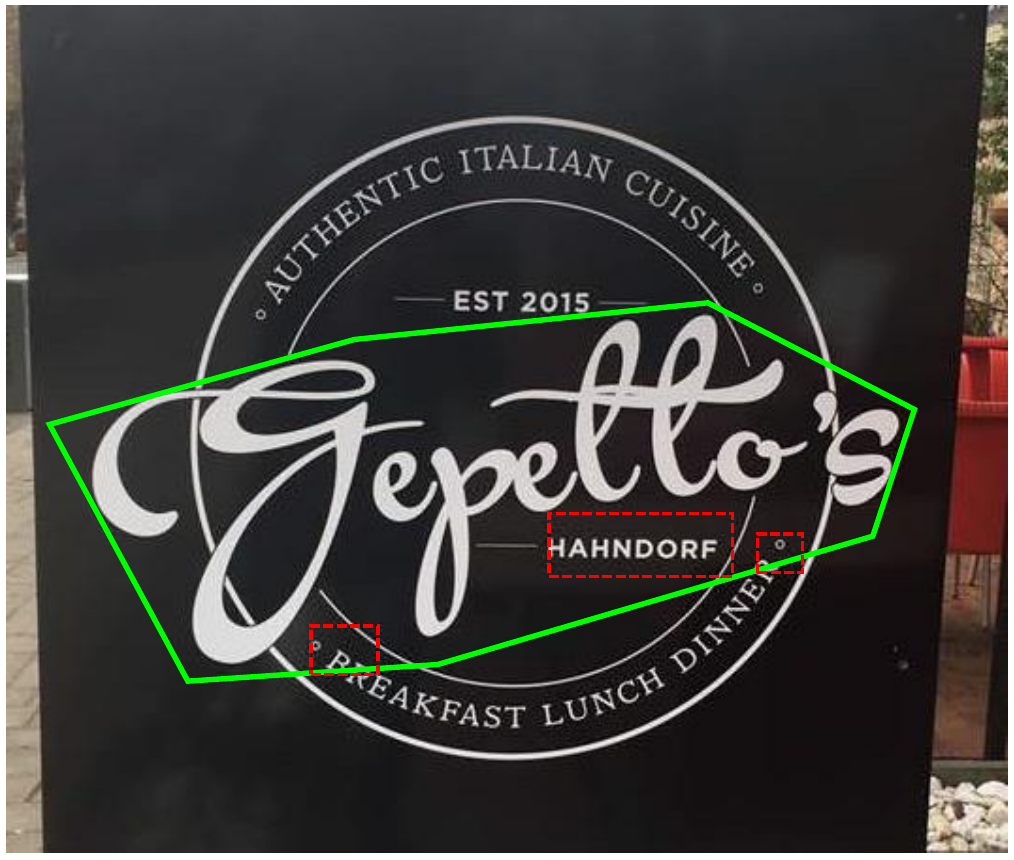}
        \caption{Polygon (172s).}
        \label{fig:tgt_poly} 
    \end{subfigure}
    \begin{subfigure}{0.18\linewidth}
        \includegraphics[width=\linewidth]{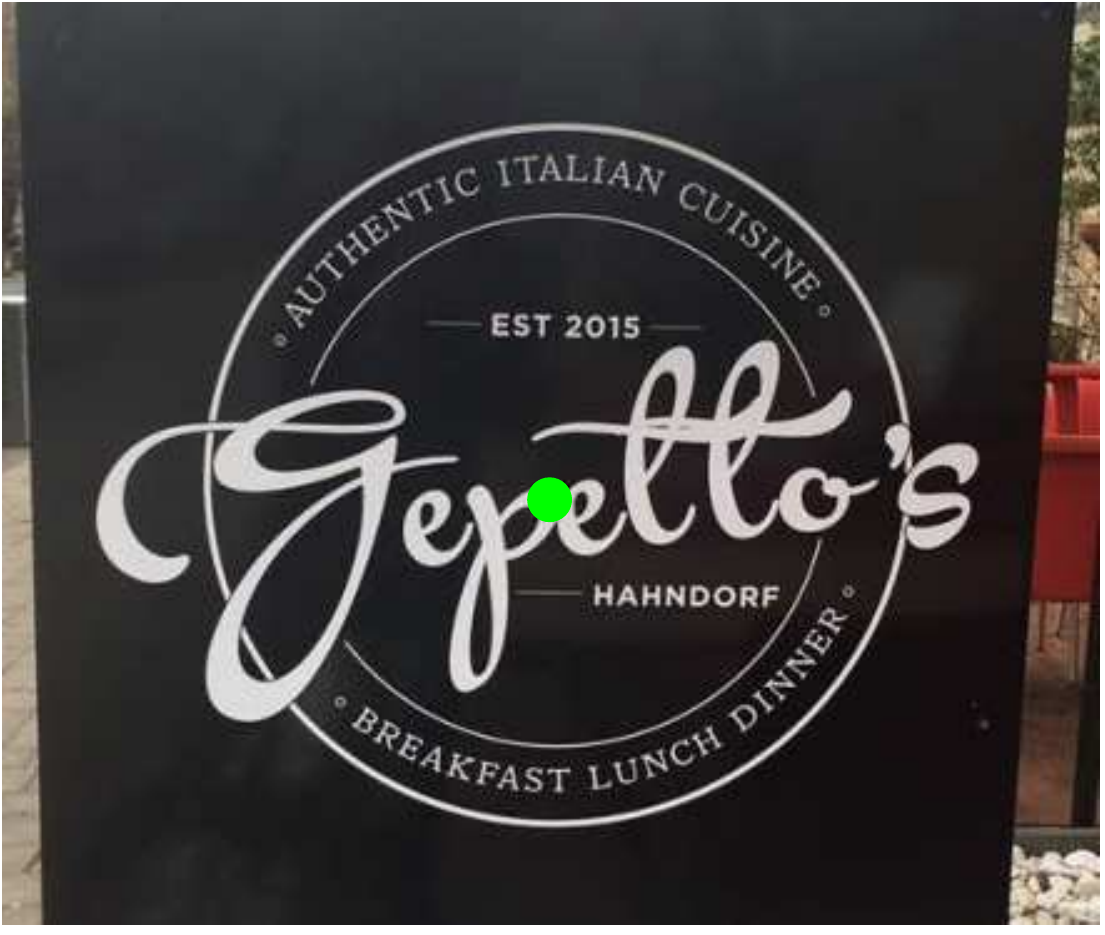}
        \caption{Single-Point (11s).}
        \label{fig:tgt_point} 
    \end{subfigure}
    \caption{Different annotation styles and their time cost (for all the text instances in the sample image) measured by the LabelMe\protect\footnotemark  tool. Green areas are positive samples, while red dashed boxes are noises that may be possibly included. The time cost of single-point annotation is more than 50 times faster than character-level annotation.}
    \label{fig:trn_tgt}
\end{figure*}

\begin{figure}[t!]
\centering
    \begin{subfigure}{0.325\linewidth}
        \includegraphics[width=2.4cm, height=1.6cm]{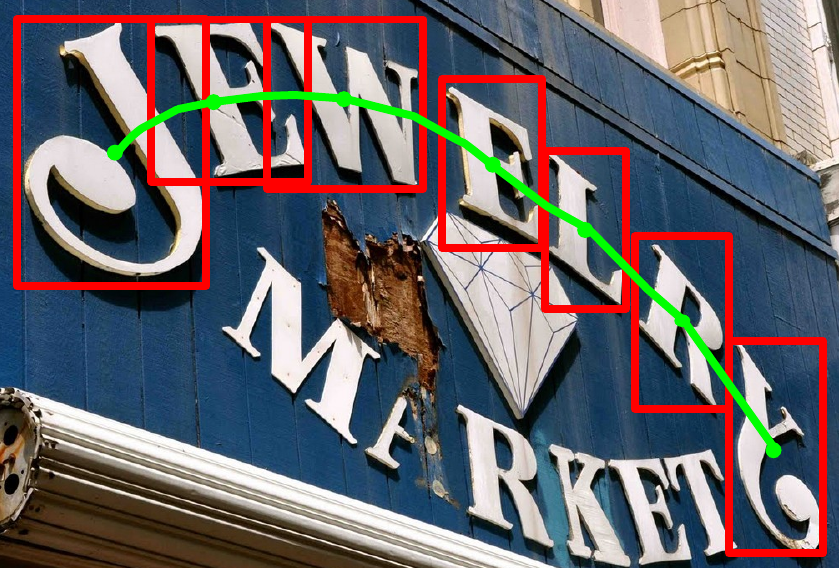}
        \caption{Text Dragon~\cite{feng2019textdragon}}
        \label{fig:textdragon}
    \end{subfigure}
    \begin{subfigure}{0.325\linewidth}
        \includegraphics[width=2.4cm, height=1.6cm]{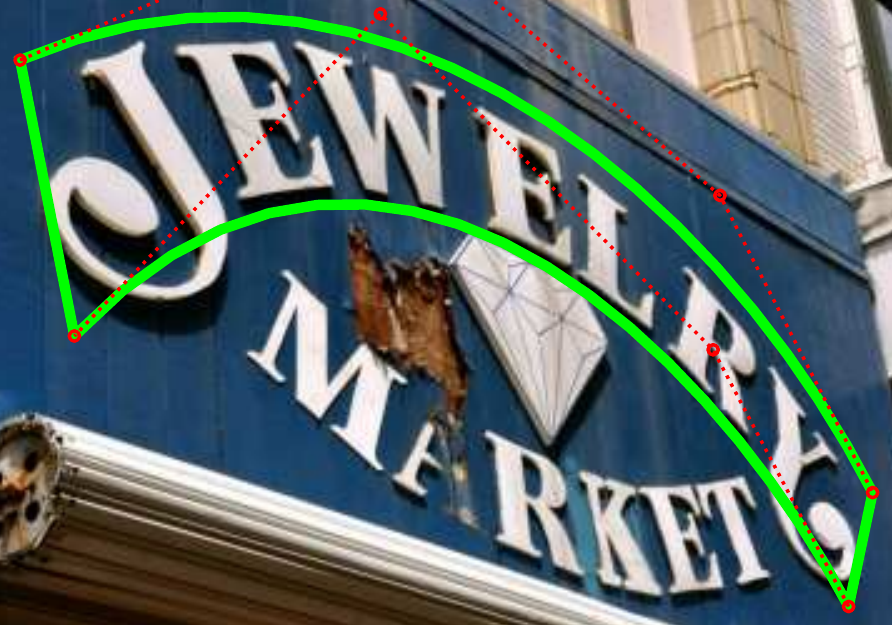}
        \caption{ABCNet~\cite{liu2020abcnet}}
        \label{fig:bzcurve}
    \end{subfigure}
    \begin{subfigure}{0.325\linewidth}
        \includegraphics[width=2.4cm, height=1.6cm]{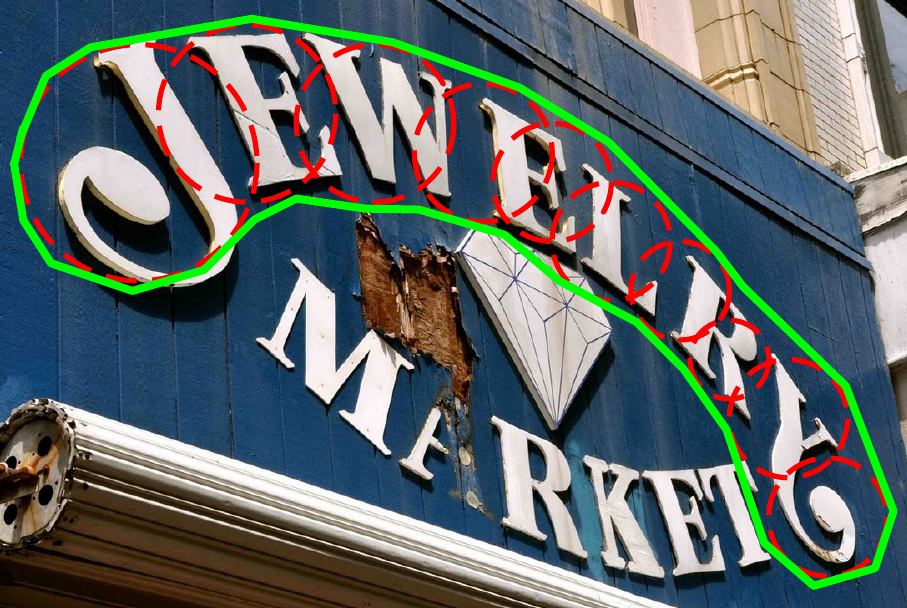}
        \caption{Text Snake~\cite{long2018textsnake}}
        \label{fig:textsnake}
    \end{subfigure}
    \caption{Some recent representations of text instances.}
    \label{fig:novel_representations}
\end{figure}

The focus of research in the OCR community has moved on from horizontal~\cite{tian2016detecting, liao2017textboxes} and multi-oriented text~\cite{zhou2017east, yao2012detecting,liu2017deep} to arbitrarily shaped text~\cite{liu2020abcnet, lyu2018mask} in recent years, accompanied by the annotation format from horizontal rectangles, to quadrilaterals, and to polygons. The fact that regular bounding boxes are prone to involve noises has been well studied in previous works (see Fig.~\ref{fig:trn_tgt}), which has proved that character-level and polygonal annotations can effectively lift the model performance~\cite{lyu2018mask, liao2020masktext, xing2019convolutional}. Furthermore, many efforts have been made to develop more sophisticated representations to fit arbitrarily shaped text instances~\cite{feng2019textdragon, wang2020textray, liu2020abcnet, zhu2021fourier, long2018textsnake} (see Fig.~\ref{fig:novel_representations}). For example, Text Dragon~\cite{feng2019textdragon} utilizes character-level bounding boxes to generate centerlines for enabling the prediction of local geometry attributes, ABCNet~\cite{liu2020abcnet} converts polygon annotations to Bezier curves for representing curved text instances, and Text Snake~\cite{long2018textsnake} describes text instances by a series of ordered disks centered at symmetric axes. However, these novel representations are primarily and carefully designed by experts based on prior knowledge, heavily relying on highly customized network architecture (\emph{e.g.}, specified Region of Interest (RoI) modules) and consuming more expensive annotations (\emph{e.g.}, character-level annotations), limiting their generalization ability for practical applications. 

To reduce the cost of data labeling, some researchers~\cite{tian2017wetext, bartz2018see, hu2017wordsup, baek2019character} have explored training the OCR models with coarse annotations in a weakly-supervised manner. These methods can mainly be separated into two categories, \emph{i.e.}, (1) bootstrapping labels to finer granularity and (2) training with partial annotations. The former usually derives character-level labels from word- or line-level annotations; thus, the models could enjoy the well-understood advantage of character-level supervision without introducing overhead costs. The latter is committed to achieving competitive performance with fewer training samples. However, both methods still rely on the costly bounding box annotations.

One of the underlying problems that prevent replacing the bounding box with a simpler annotation format, such as a single-point, is that most text spotters rely on RoI-like sampling strategies to extract the shared backbone features. For example, Mask TextSpotter requires mask prediction inside a RoI~\cite{liao2020masktext}; ABCNet~\cite{liu2020abcnet} proposes BezeirAlign, while TextDragon~\cite{feng2019textdragon} introduces RoISlide to unify the detection and recognition heads. In this paper, inspired by the recent success of a sequence-based object detector Pix2Seq~\cite{chen2021pix2seq}, we show that the text spotter can be trained with a single-point, also termed as the indicated point (see Fig.~\ref{fig:tgt_point}). Thanks to such a concise form of annotation, labeling time can be significantly saved, \emph{e.g.}, it only takes less than one-fiftieth of the time to label single-points for the sample image shown in Fig.~\ref{fig:trn_tgt} compared with annotating character-level bounding boxes, which is extremely tortuous especially for the small and vague text instances. Another motivating factor in selecting point annotation is that a clean and efficient OCR pipeline can be developed, discarding the complex post-processing module and sampling strategies; thus, the ambiguity introduced by RoIs (see red dashed regions in Fig.~\ref{fig:trn_tgt}) can be alleviated.
To the best of our knowledge, this is the first attempt to simplify the bounding box to a single-point supervision signal in the OCR community. The main contributions of this work are summarized as follows:
\begin{itemize}
    \item For the first time, we show that the text spotters can be supervised by a simple yet effective single-point representation. Such a straightforward annotation paradigm can considerably reduce the labeling costs, making it possible to access large-scale OCR data in the future.
    \item We propose a new Transformer-based~\cite{vaswani2017attention} scene text spotter, which forms the text spotting as a language modeling task. Given an input image, our method predicts a discrete token sequence that includes both detection and recognition results. Benefiting from such a concise pipeline, the complex post-processing and sampling strategies designed based on prior knowledge can be discarded, showing great potential in terms of flexibility and generality.
    \item To evaluate the effectiveness of proposed methods, extensive experiments and ablations are conducted on four widely used OCR datasets, \emph{i.e.}, ICDAR 2013~\cite{karatzas2013icdar}, ICDAR 2015~\cite{karatzas2015icdar}, Total-Text~\cite{ch2017total}, and SCUT-CTW1500~\cite{liu2019curved}, involving both horizontal and arbitrarily shaped texts. The results show that the proposed \methodName\ can achieve state-of-the-art performance compared with existing approaches.
\end{itemize}

\footnotetext{\sf  https://github.com/wkentaro/labelme}

\begin{figure*}[t!]
	\centering
	\includegraphics[width=0.95\linewidth]{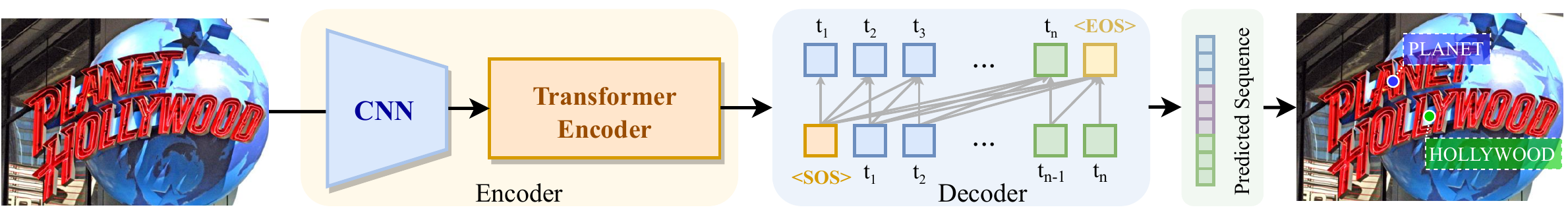}
	\caption{Overall framework of the proposed \methodName. The visual and contextual features are first extracted by a series of CNN and Transformer encoders. Then, the features are auto-regressively decoded into a sequence that contains both localization and recognition information, which is subsequently translated into point coordinates and text transcriptions. Only a point-level annotation is required for training.}
	\label{fig:method}
\end{figure*}

\subsection{Related Work}

In the past decades, a variety of scene text datasets using different annotation styles have been proposed, focusing on various scenarios, including horizontal text~\cite{karatzas2013icdar, karatzas2011icdar} described by rectangles (Fig.~\ref{fig:tgt_rect}), multi-oriented text~\cite{karatzas2015icdar, nayef2019icdar2019} represented by quadrilaterals (Fig.~\ref{fig:tgt_quad}), and arbitrarily shaped text~\cite{ch2017total, liu2019curved, chng2019icdar2019} labeled by polygons (Fig.~\ref{fig:tgt_poly}). These forms of annotations have facilitated the development of corresponding OCR algorithms. For example, earlier works~\cite{li2017towards} usually adapt the generic object detectors to scene text spotting, where feature maps are shared between detection and recognition heads via RoI modules. These approaches follow the sampling mechanism designed for generic object detection and utilize rectangles to express text instances, thus performing worse on non-horizontal targets. Later, some methods~\cite{busta2017deep, he2018end, liu2018fots} replace the rectangular bounding boxes with quadrilaterals by modifying the regular Region Proposal Network (RPN) to generate oriented proposals, enabling better performance for multi-oriented texts. Recently, with the presentation of the curved scene text datasets~\cite{ch2017total, liu2019curved, chng2019icdar2019}, the research interest of the OCR community has shifted to more challenging arbitrarily shaped texts. Generally, there are two widely adopted solutions to solve the arbitrarily shaped text spotting task, \emph{i.e.}, segmentation-based~\cite{liao2020masktext, qiao2021mango, wang2021pan++, qin2019towards} and regression-based methods~\cite{liu2020abcnet, boundary2020, feng2019textdragon}. The former first predicts masks to segment text instances, then the features inside text regions are sampled and grouped for further recognition. For example, Mask TextSpotterv3~\cite{liao2020masktext} proposes a Segmentation Proposal Network (SPN) instead of the regular RPN to decouple neighboring text instances accurately, thus significantly improving the performance. In addition, regression-based methods usually parameterize the text instances as a sequence of coordinates and subsequently learn to predict them. For instance, ABCNet~\cite{liu2020abcnet} converts polygons into Bezier curves, significantly improving performance on curved scene texts. Wang et al.~\cite{boundary2020} first localize the boundary points of text instances, then the features rectified by Thin-Plate-Spline are fed into the recognition branch, demonstrating promising accuracy on arbitrary-shaped instances. Moreover, Xing et al.~\cite{xing2019convolutional} boost the text spotting performance by utilizing character-level annotations, where character bounding boxes, as well as type segmentation maps, are predicted simultaneously, enabling impressive performance. 
Even though different representations are adopted in the above methods to describe the text instances, they are all actually derived from one of the rectangular, quadrilateral, or polygonal bounding boxes. Such annotations must be carefully labeled by human beings, thus are quite expensive, limiting the scale of training datasets.

In this paper, we propose Single-Point Text Spotting (\methodName), which is, to the best of our knowledge, the first scene text spotter that does not rely on bounding box annotations at all. Specifically, each text instance is represented by a single-point (see Fig.~\ref{fig:tgt_point}) with a meager cost. The fact that this point does not need to be accurately marked further demonstrates the possibility of learning in a weakly supervised manner, considerably lowering down the labeling cost.

\section{Methodology}\label{sec:method}
Most of the existing text spotting algorithms treat the problem as two sub-tasks, \emph{i.e.}, text detection and recognition, albeit the entire network might be end-to-end optimized. Customized modules such as BezierAlign~\cite{liu2020abcnet}, RoISlide~\cite{feng2019textdragon}, and RoIMasking~\cite{liao2020masktext} are required to bridge the detection and recognition modules, where backbone features are cropped and shared between detection and recognition heads. Under such types of design, the recognition and detection modules are highly coupled. For example, the features fed to the recognition head are usually cropped from the ground-truth bounding box at the training stage since  detection results are not good enough in the first iterations; thus, the recognition result is susceptible to interference from the detected bounding box during the test phase.

Recently, Pix2Seq~\cite{chen2021pix2seq} pioneered to cast the generic object detection problem as a language modeling task, based on an intuitive assumption that if a deep model knows what and where the target is, it can be taught to tell the results by the desired sequence. Thanks to the concise pipeline, labels with different attributes such as location coordinates and object categories can be integrated into a single sequence, enabling an end-to-end trainable framework without task-specific modules (\emph{e.g.}, Region Proposal Networks and RoI pooling layers), which can thus be adapted to the text spotting task. Inspired by this, we propose Single-Point Text Spotting (\methodName). Unlike Pix2Seq which is designed for object detection only and still requires the bounding boxes for all instances, our SPTS tackles both text detection and recognition as a end-to-end sequence prediction task, using the single point location and text annotations. Compared with existing text spotting approaches, SPTS follows a much more simple and concise pipeline where the input images are translated into a sequence containing location and recognition results, genuinely enabling the text detection and recognition task simultaneously.

Specifically, as shown in Fig.~\ref{fig:method}, each input image is first sequentially encoded by a CNN and a Transformer encoder~\cite{vaswani2017attention} to extract visual and contextual features. Then, the captured features are decoded by a Transformer decoder~\cite{vaswani2017attention}, where tokens are predicted in an auto-regressive manner. Unlike previous algorithms, we further simplify the bounding box to the center of the text instance, the corner point located at the top left of the first character, or the random point within the text instance, as described in Fig.~\ref{fig:pts_pos}. Benefiting from such a simple yet effective representation, the modules carefully designed based on prior knowledge, such as grouping strategies utilized in segmentation-based methods and feature sampling blocks equipped in box-based text spotters, can be eschewed. Therefore, the recognition accuracy will not be limited by poor detection results, significantly improving the model robustness.

\subsection{Sequence Construction}
\label{sec:seq construction}
\begin{figure}[t!]
    \centering
    \includegraphics[width=0.9\columnwidth]{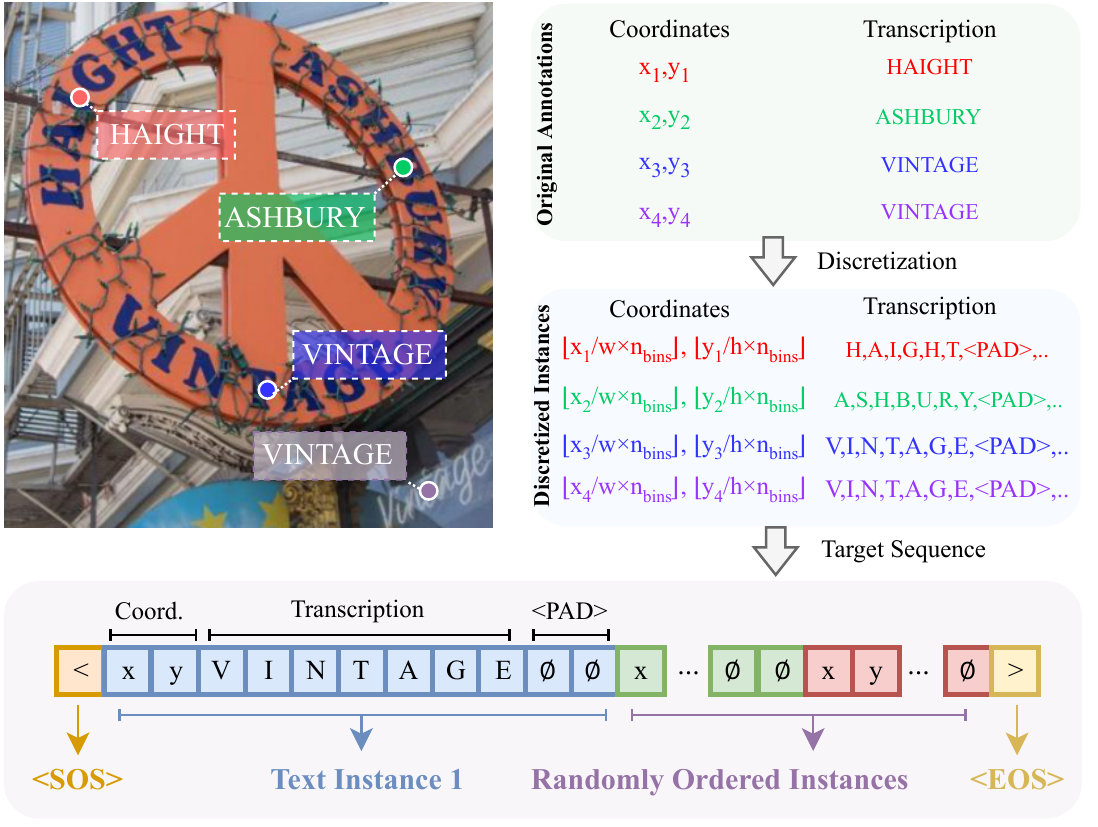}
    \caption{Pipeline of the sequence construction.}
    \label{fig:seq_constrcut}
\end{figure}

The fact that a sequence can carry information with multiple attributes naturally enables the text spotting task, where text instances are simultaneously localized and recognized. To express the target text instances by a sequence, it is required to convert the continuous descriptions (\emph{e.g.}, bounding boxes) to a discretized space. To this end, as shown in Fig.~\ref{fig:seq_constrcut}, we follow Pix2Seq~\cite{chen2021pix2seq} to build the target sequence; what distinguishes our methods is that we further simplify the bounding box to a single-point and use the variable-length transcription instead of the single-token object category. 

Specifically, the continuous coordinates of the central point of the text instance are uniformly discretized into integers between $[1, n_{bins}]$, where $n_{bins}$ controls the degree of discretization. For example, an image with a long side of 800 pixels requires only $n_{bins}=800$ to achieve zero quantization error. Note that the central point of the text instance is obtained by averaging the upper and lower midpoints as shown in Fig. \ref{fig:ctr}. As so far, a text instance can thereby be represented by a sequence of three parts, \emph{i.e.}, $[x, y, t]$, where $(x, y)$ are the discretized coordinates and $t$ is the transcription text. Notably, the transcriptions are inherently discrete, \emph{i.e.}, each of the characters represents a category, thus can be easily appended to the sequence. However, different from the generic object detection that has a relatively fixed vocabulary (each $t$ represents an object category, such as pedestrian), $t$ can be any natural language text of any length in our task, resulting in a variable length of the target sequence, which may further cause misalignment issues and can consume more computational resources. To eliminate such problems, we pad or truncate the texts to a fixed length $l_{tr}$, where the $<$PAD$>$ token is used to fill the vacancy for shorter text instances. In addition, like other language modeling methods, $<$SOS$>$ and $<$EOS$>$ tokens are inserted to the head and tail of the sequence, indicating the start and the end of a sequence, respectively. Therefore, given an image that contains $n_{ti}$ text instances, the constructed sequence will include $(2+l_{tr})\times n_{ti}$ discrete tokens, where the text instances would be randomly ordered, following previous works~\cite{chen2021pix2seq}. Supposing there are $n_{cls}$ categories of characters (\emph{e.g.}, 97 for English characters and symbols), the vocabulary size of the dictionary used to tokenize the sequence can be calculated as $n_{bins} + n_{cls} + 3$, where the extra three classes are for $<$PAD$>$, $<$SOS$>$, and $<$EOS$>$ tokens. Empirically, we set the $l_{tr}$ and $n_{bins}$ to 25 and 1,000, respectively, in our experiments. Moreover, the maximum value of $n_{ti}$ is set to 60, which means the sequence containing more than 60 text instances will be truncated.

\begin{figure}[t!]
    \centering
    \includegraphics[width=0.9\linewidth]{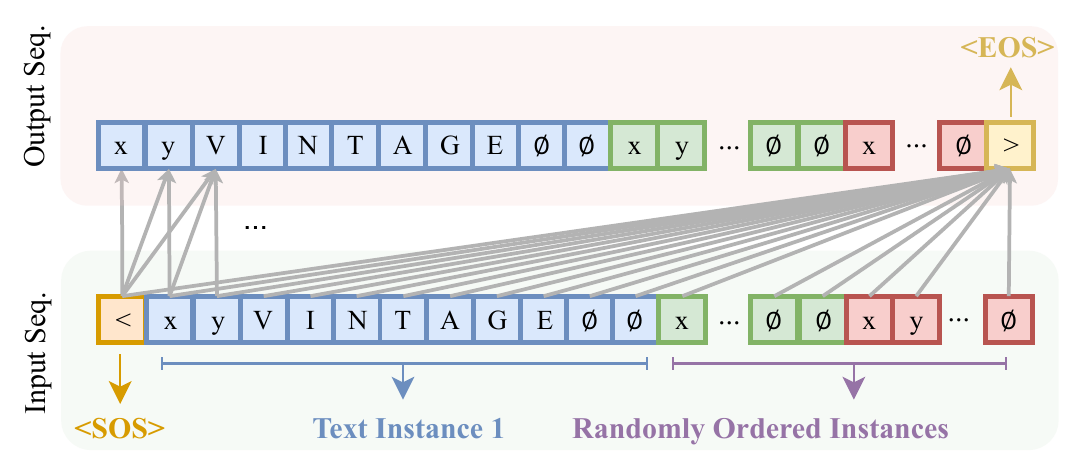}
    \caption{Input and output sequences of the decoder.}
    \label{fig:input_output_seq}
\end{figure}

\subsection{Model Training}



\def\bw{{\bf w}}
\def\bs{{\bf s}}
\def\bI{{\bf I}}
%
Based on the constructed sequence, the input and output sequences of the Transformer decoder are shown in Fig.~\ref{fig:input_output_seq}. Since the \methodName\ is trained to predict tokens, it only requires to maximize the likelihood loss at training time, which can be written as:
\begin{equation}
\label{eq_objective}
%
%
{\rm maximize} \sum_{i=1}^{L} {\bw}_i \log
 P(\tilde{{\bs}}_i | {\bI}, {\bs}_{1:i}),
\end{equation}
where $\textbf{I}$ is the input image, $\tilde{\textbf{s}}$ is the output sequence, $\textbf{s}$ is the input sequence, $ L$ is the length of the sequence, and $\textbf{w}_i$ is the weight of the likelihood of the $i$-th token, which is empirically set to 1.

\subsection{Inference}

At the inference stage, SPTS auto-regressively predicts the tokens until the end of the sequence token $<$EOS$>$ occurs. The predicted sequence will subsequently be divided into multiple segments, each of which contains $2 + l_{tr}$ tokens. Then, the tokens can be easily translated into the point coordinates and transcriptions, yielding the text spotting results. In addition, the likelihood of all tokens in the corresponding segment is averaged and assigned as a confidence score to filter the original outputs, which effectively removes redundant and false-positive predictions.

\section{Experiments}\label{sec:exp}

We report the experimental results on four widely used benchmarks including horizontal dataset ICDAR 2013~\cite{karatzas2013icdar}, multi-oriented dataset ICDAR 2015~\cite{karatzas2015icdar}, and arbitrarily shaped datasets Total-Text~\cite{ch2017total} and SCUT-CTW1500~\cite{liu2019curved}. 
    
\subsection{Datasets} 
    
{\bf Curved Synthetic Dataset 150k.} 
It is admitted that the performance of text spotters can be improved by pre-training on synthesized samples. Following previous work~\cite{liu2020abcnet}, we use the 150k synthetic images generated by the SynthText~\cite{gupta2016synthetic} toolbox, which contains around one-third of curved texts and two-third of horizontal instances.

{\bf ICDAR 2013}~\cite{karatzas2013icdar} 
contains 229 training and 233 testing samples, while the images are primarily captured in a controlled environment, where the text contents of interest are explicitly focused in horizontal.

{\bf ICDAR 2015}~\cite{karatzas2015icdar} 
consists of 1,000 training and 500 testing images that were incidentally captured, containing multi-oriented text instances presented in complicated backgrounds with strong variations in blur, distortions, etc.

{\bf Total-Text}~\cite{ch2017total} 
includes 1,255 training and 300 testing images, where at least one curved sample is presented in each image and annotated with polygonal bounding box at the word-level.
    
{\bf SCUT-CTW1500}~\cite{liu2019curved} 
is another widely used benchmark designed for spotting arbitrary shaped scene text, involving 1,000 and 500 images for training and testing, respectively. The text instances are labeled by polygons at text-line level.
    
    
\subsection{Evaluation Protocol}
\label{subsec:eval_protocol}

\begin{figure}[t!]
    \centering
    \includegraphics[width=0.8\columnwidth]{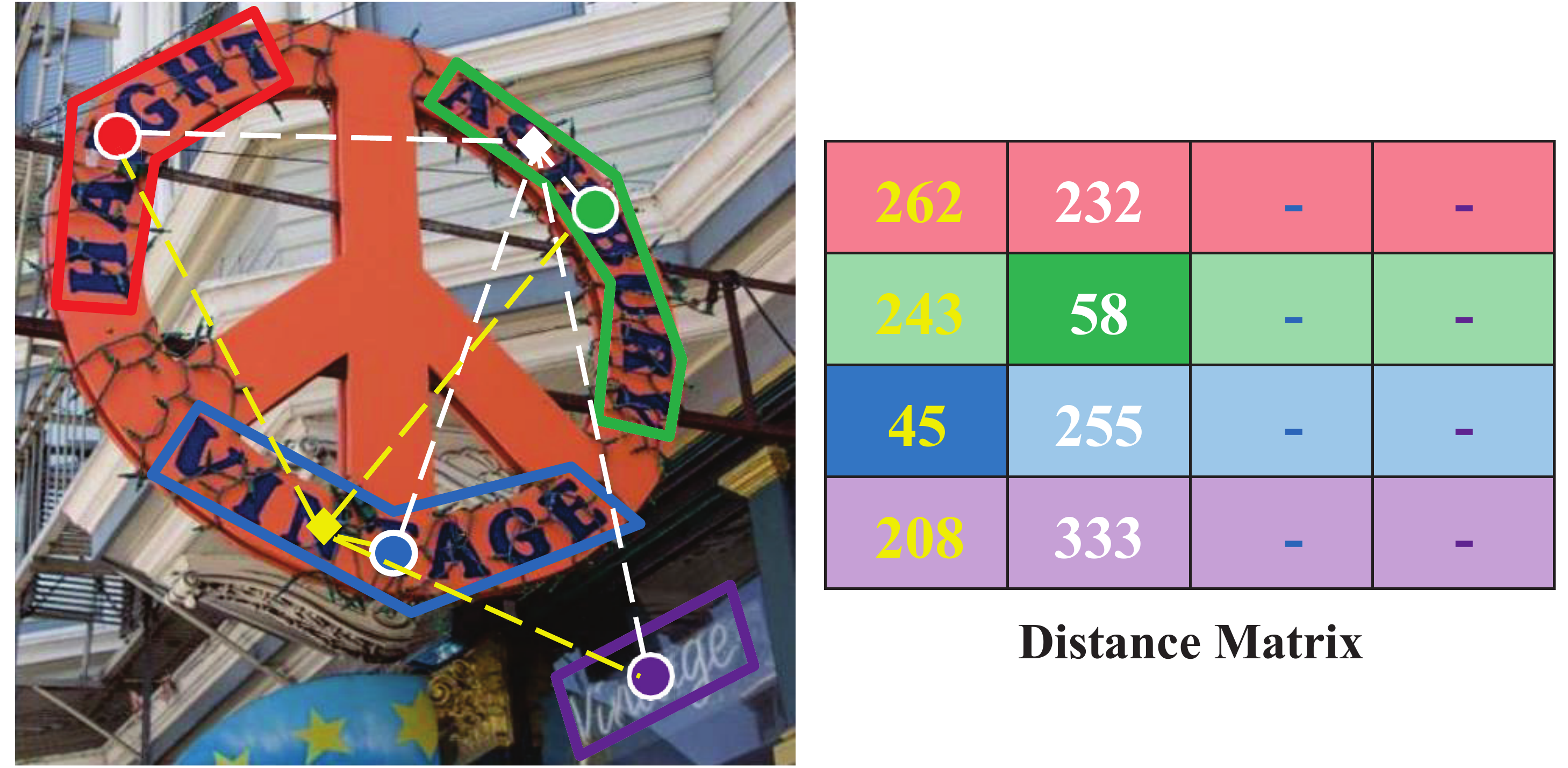}
    \caption{Illustration of the point-based evaluation metric. Diamonds are predicted points and circles represent ground-truth.}
    \label{fig:eval}
\end{figure}

The existing evaluation protocol of text spotting tasks consists of two steps. Firstly, the intersection over union (IoU) scores between ground-truth (GT) and detected boxes are calculated; and only if the IoU score is larger than a designated threshold (usually set to 0.5), the boxes are matched. Then, the recognized content inside each matched bounding box is compared with the GT transcription; only if the predicted text is the same as the GT will it contribute to the end-to-end accuracy. However, in the proposed method, each text instance is represented by a single-point; thus, the evaluation metric based on the IoU is not available to measure the performance. Meanwhile, comparing the localization performance between bounding-box-based methods and the proposed point-based \methodName\ might be unfair, \emph{e.g.}, directly treating points inside a bounding box as true positives may overestimate the detection performance. To this end, we propose a new evaluation metric to ensure a relatively fair comparison to existing approaches, which mainly considers the end-to-end accuracy as it reflects both detection and recognition performance (failure detections usually lead to incorrect recognition results). Specifically, as shown in Fig.~\ref{fig:eval}, we modified the text instance matching rule by replacing the IoU metric with a distance metric, \emph{i.e.}, the predicted point that has the nearest distance to the central point of the GT box would be selected, and the recognition results will be measured by the same full-matching rules used in existing benchmarks. Only one predicted point with the highest confidence would be matched to the ground truth; others are then marked as false positives. 
    
To explore whether the proposed evaluation protocol can genuinely represent the model accuracy, Table~\ref{tab:ab_metric} compares the end-to-end recognition accuracy of ABCNetv1~\cite{liu2020abcnet} and ABCNetv2~\cite{liu2021abcnetv2} on Total-Text~\cite{ch2017total} and SCUT-CTW1500~\cite{liu2019curved} under two metrics, \emph{i.e.}, the commonly used bounding box metric that is based on IoU, and the proposed point-based metric. The results demonstrate that the point-based evaluation protocol can well reflect the performance, where the difference between the values evaluated by box-based and point-based metrics are 
no more than 0.5\%. For example, the ABCNetv1 model achieves 53.5\% and 53.0\% scores on the SCUT-CTW1500 dataset under the two metrics, respectively. Therefore, we use the point-based metric to evaluate the proposed \methodName\ in the following experiments.
    
\begin{table}[t!]
    \centering
    \caption{Comparison of the end-to-end recognition performance evaluated by the proposed point-based metric and box-based metric. Results are reproduced using official codes.}
    \label{tab:ab_metric}
    \footnotesize
    \begin{tabular}{r|cc|cc}\hline
        \multirow{2}{*}{Method} & \multicolumn{2}{c|}{Total-Text} & \multicolumn{2}{c}{SCUT-CTW1500} \\ \cline{2-5}
         & \multicolumn{1}{c|}{Box} & Point & \multicolumn{1}{c|}{Box} & \multicolumn{1}{c}{Point} \\ \hline
        ABCNetv1~\cite{liu2020abcnet} & 67.2 & 67.4 & 53.5 & 53.0  \\
        ABCNetv2~\cite{liu2021abcnetv2} & 71.7 & 71.9 & 57.6 & 57.1 \\ \hline
    \end{tabular}
\end{table}

\subsection{Implemented Details}
The model is first pretrained on a combination dataset that includes Curved Synthetic Dataset 150k~\cite{liu2020abcnet}, MLT-2017~\cite{nayef2017icdar2017}, ICDAR 2013~\cite{karatzas2013icdar}, ICDAR 2015~\cite{karatzas2015icdar}, and Total-Text~\cite{ch2017total} for 150 epochs, which is optimized by the AdamW~\cite{loshchilov2017decoupled} with an initial learning rate of $5\times 10^{-4}$, while the learning rate is linearly decayed to $1\times 10^{-5}$. After pretraining, the model is then fine-tuned on the training split of each target dataset for another 200 epochs, with a fixed learning rate of $1\times 10^{-5}$. The entire model is distributively trained on 32 NVIDIA V100 GPUs with a batch size of 32. Note that the effective batch size is 64 because two independent augmentations are performed on each image in a mini-batch, following \cite{chen2021pix2seq,hoffer2020augment}. In addition, we utilize ResNet-50 as the backbone network, while both the Transformer encoder and decoder consist of 6 layers with eight heads. Regarding the architecture of the Transformer, we adopt the Pre-LN Transformer~\cite{xiong2020layer}.
During training, short size of the input image is randomly resized to a range from 640 to 896 (intervals of 32) while keeping the longer side shorter than 1,600 pixels. Random cropping and rotating are employed for data augmentation. At the inference stage, we resize short edge to 1,000 while keeping longer side shorter than 1824 pixels, following the previous works~\cite{liu2020abcnet,liu2021abcnetv2}.
    
\subsection{Ablation Study} 
\label{subsec:abs}

\subsubsection{Ablation study of the position of the indicated point}
\begin{figure}[t!]
    \centering
    \begin{subfigure}{0.3\linewidth}
        \includegraphics[width=2.4cm, height=1.6cm]{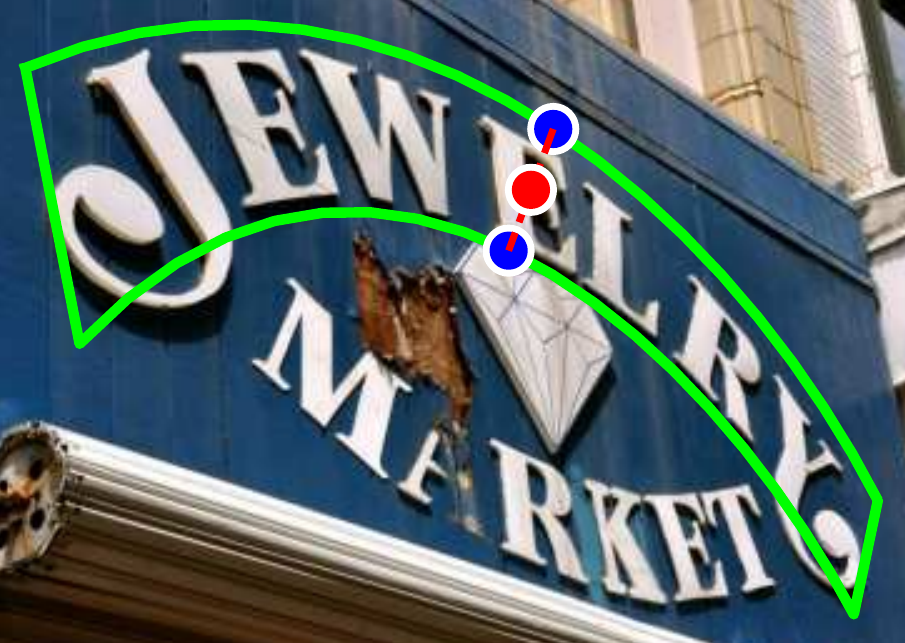}
        \caption{Central}
        \label{fig:ctr}
    \end{subfigure}
    \begin{subfigure}{0.3\linewidth}
        \includegraphics[width=2.4cm, height=1.6cm]{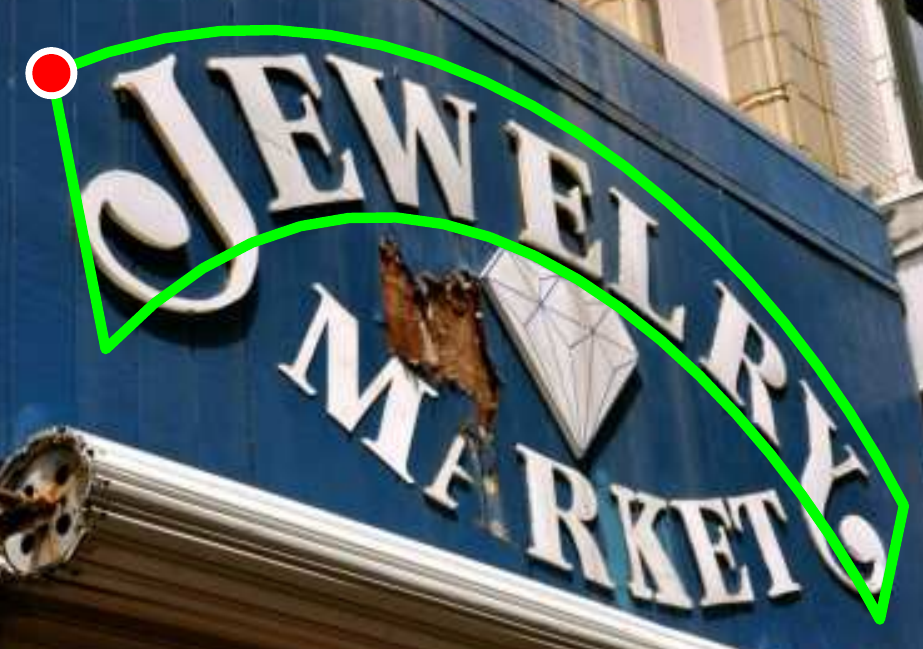}
        \caption{Top-left}
        \label{fig:tl}
    \end{subfigure}
    \begin{subfigure}{0.3\linewidth}
        \includegraphics[width=2.4cm, height=1.6cm]{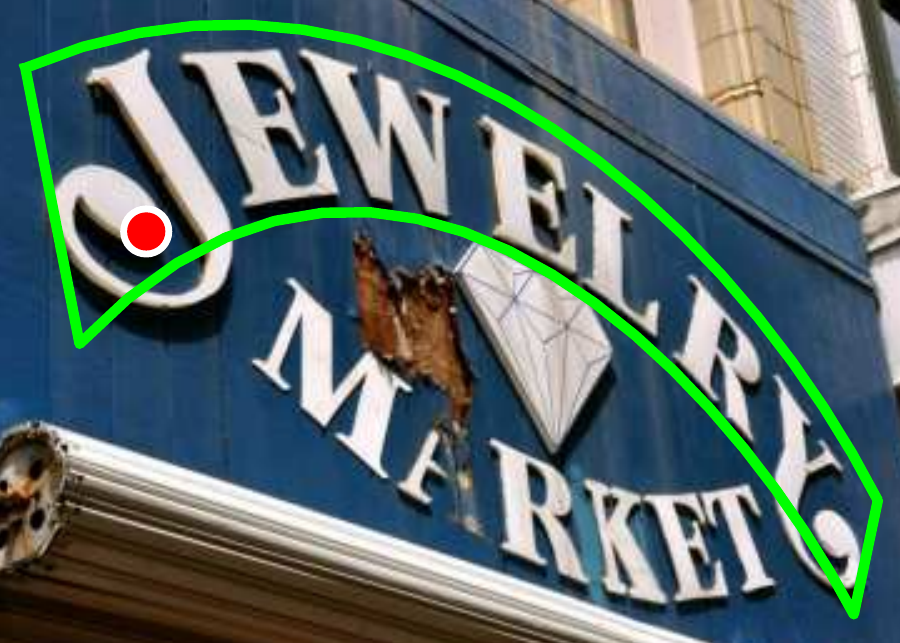}
        \caption{Random}
        \label{fig:rand}
    \end{subfigure}
    \caption{Indicated points (red) using different positions.}
    \label{fig:pts_pos}
\end{figure}

In this paper, we propose to simplify the bounding box to a single-point. Intuitively, all points in the region enclosed by the bounding box should be able to represent the target text instance. To explore the differences, we conduct ablation studies that use three different strategies to get the indicated points (see Fig.~\ref{fig:pts_pos}), \emph{i.e.}, the \emph{central} point obtained by averaging the upper and lower midpoints, the \emph{top-left} corner, and the \emph{random} point inside the box. It should be noted that, we use the corresponding ground-truth here to calculate the distance matrix for evaluating the performance, \emph{i.e.}, the distance to the ground-truth top-left point is used for \emph{top-left}, the distance to the ground-truth central point for \emph{central}, and the closest distance to the ground-truth polygon for \emph{random}.
    
\begin{table}[t!]
    \centering
    \caption{Ablation study of the position of the indicated point.}
    \label{tab:ab1_position_point}
    \footnotesize
    \begin{tabular}{c|cc|cc}\hline
    \multirow{2}{*}{Position} & \multicolumn{2}{c|}{E2E Total-Text} & \multicolumn{2}{c}{E2E CTW1500} \\ \cline{2-5} 
     & \multicolumn{1}{c|}{None} & Full & \multicolumn{1}{c|}{None} & \multicolumn{1}{c}{Full} \\ \hline
    Central & \textbf{74.2} & \textbf{82.4} & \textbf{63.6} & \textbf{83.8} \\
    Top-left & 71.6 & 79.7 & 61.4 & 82.0 \\
    Random & 73.2 & 80.8 & 62.3 & 81.1 \\ \hline
    \end{tabular}
\end{table}

The results are shown in Table \ref{tab:ab1_position_point}, where the results of \emph{central}, \emph{top-left}, and \emph{random} are close on both datasets. It suggests that the performance is not very sensitive to the positions of the point annotation.

\begin{table}[t!]
\centering
\caption{Comparison with different shapes of bounding boxes. $N_{p}$ is the number of parameters required to describe the location of text instances by different representations.}
\label{tab:ab_box_shape}
\footnotesize
\begin{tabular}{r|cc|cc|c}
\hline
\multirow{2}{*}{Variants} & \multicolumn{2}{c|}{Total-Text}                   & \multicolumn{2}{c|}{SCUT-CTW1500} &
\multirow{2}{*}{$N_{p}$} \\ \cline{2-5} 
                        & \multicolumn{1}{c|}{None}    & \multicolumn{1}{c|}{Full} 
                        & \multicolumn{1}{c|}{None}    & \multicolumn{1}{c|}{Full}
                        & \\ \hline
SPTS-Bezier& 60.6 & 71.6 & 52.6 & 73.9 & 16 \\ 
SPTS-Rect & 71.6 & 80.4 & 62.2 & 82.3 & 4 \\  
SPTS-Point &\textbf{74.2} & \textbf{82.4} & \textbf{63.6} & \textbf{83.8} & 2  \\ \hline
\end{tabular}
\end{table}

\begin{table}[t!]
    \centering
    \caption{End-to-end recognition results on ICDAR 2013. ``S'', ``W'', and ``G'' represent recognition with “Strong”, “Weak”, and “Generic” lexicon, respectively.}
    \label{ICDAR 2013 End-to-End recognition result}
    \footnotesize
    \begin{tabular}{r|c|c|c}
    \hline
    \multirow{2}{*}{Method} & \multicolumn{3}{c}{IC13 End-to-End} \\ \cline{2-4} 
                            & \multicolumn{1}{c|}{S}    & \multicolumn{1}{c|}{W}    & G    \\ \hline
    \multicolumn{4}{c}{Bounding Box-based methods} \\ \hline                        
    Jaderberg et al.\  \cite{jaderberg2016reading} & 86.4 & - & -\\ 
    Textboxes \cite{liao2017textboxes} & 91.6 & 89.7 & 83.9\\ 
    Deep Text Spotter \cite{busta2017deep} & 89.0 & 86.0 & 77.0\\ 
    Li et al. \cite{li2017towards} & 91.1 & 89.8 & 84.6 \\ 
    MaskTextSpotter \cite{lyu2018mask} & \underline{92.2} & \underline{91.1} & \underline{86.5} \\ 
    \hline 
    \multicolumn{4}{c}{Point-based method} \\ \hline
    \methodName\ (Ours) & \textbf{93.3} & \textbf{91.7} & \textbf{88.5} \\ \hline
    \end{tabular}
\end{table}
    
\begin{table}[t!]
    \centering
    \caption{End-to-end recognition results on ICDAR 2015. “S”, “W”, and “G” represent recognition with “Strong”, “Weak”, and “Generic” lexicon, respectively.}
    \label{ICDAR 2015 End-to-End recognition result}
    \footnotesize
    \begin{tabular}{r|c|c|c}
    \hline
    \multirow{2}{*}{Method} & \multicolumn{3}{c}{IC15 End-to-End}                    \\ \cline{2-4} 
                            & \multicolumn{1}{c|}{S}    & \multicolumn{1}{c|}{W}    & G    \\ \hline
    \multicolumn{4}{c}{Bounding Box-based methods} \\ \hline                               
    FOTS \cite{liu2018fots}                    & \multicolumn{1}{c|}{81.1} & \multicolumn{1}{c|}{75.9} & 60.8 \\
    Mask TextSpotter \cite{liao2019mask}        & \multicolumn{1}{c|}{83.0} & \multicolumn{1}{c|}{77.7} & \underline{73.5} \\
    CharNet \cite{xing2019convolutional}                 & \underline{83.1} & \multicolumn{1}{c|}{\textbf{79.2}} & 69.1 \\
    TextDragon \cite{feng2019textdragon} & \multicolumn{1}{c|}{82.5} & \multicolumn{1}{c|}{78.3} & 65.2 \\
    Mask TextSpotter v3 \cite{liao2020masktext} & \multicolumn{1}{c|}{\textbf{83.3}} & \multicolumn{1}{c|}{78.1} & \textbf{74.2} \\
    MANGO \cite{qiao2021mango}                   & \multicolumn{1}{c|}{81.8} & \underline{78.9} & 67.3 \\
    ABCNetV2 \cite{liu2021abcnetv2}                & \multicolumn{1}{c|}{82.7} & \multicolumn{1}{c|}{78.5} & 73.0 \\ 
    PAN++ \cite{wang2021pan++}                   & \multicolumn{1}{c|}{82.7} & \multicolumn{1}{c|}{78.2} & 69.2 \\
    \hline
    \multicolumn{4}{c}{Point-based method} \\ \hline   
    \methodName\ (Ours) &  77.5 & 70.2 & 65.8 \\ \hline
    \end{tabular}
\end{table}

\subsubsection{Comparison between different representations}
\label{sec_rect_bezier}
The proposed \methodName\ can be easily extended to produce bounding boxes by modifying the point coordinates to bounding box locations during sequence construction. Here, we conduct ablations to explore the influence by using different representations of the text instances. Specifically, three variants are explored, including the Bezier curve bounding box (\methodName-Bezier), the rectangular bounding box (SPTS-Rect), and the indicated point (\methodName-point). Since we only focus on end-to-end performance here, to minimize the impact of detection results, each method uses corresponding representations to match the GT box in the evaluation. That is to say, the single-point model (original \methodName) uses the evaluation metrics introduced in Sec.~\ref{subsec:eval_protocol}, \emph{i.e.}, distance between points; the predictions of \methodName-Rect are matched to the circumscribed rectangle of the polygonal annotations; the \methodName-Bezier adopts the original metric that matches polygon boxes. As shown in Table~\ref{tab:ab_box_shape}, the \methodName-point achieves the best performance on both the Total-Text and SCUT-CTW1500 datasets, outperforming the other two representations by a large margin. Such experimental results suggest that a low-cost annotation, \emph{i.e.}, the indicated point, is capable of providing supervision for the text spotting task. The reason for the low performance of \methodName-Bezier and \methodName-Rect may be that longer sequences (\emph{e.g.}, \methodName-Bezier with $N_{p}=16$ \emph{vs.} \methodName-Point with $N_{p}=2$) make them difficult to converge; thus, \methodName-Bezier and \methodName-Rect cannot achieve comparable accuracy under the same training schedule.

\subsection{Comparison with Existing Methods on Scene Text Benchmarks}

\begin{table}[t!]
\centering
\caption{End-to-end recognition results on Total-Text. ``None"
represents lexicon-free. ``Full" represents that we use all the words appeared in the test set.}
\label{Total-Text e2e}
\footnotesize
\begin{tabular}{r|c|c}
\hline
\multirow{2}*{Method} & \multicolumn{2}{c}{Total-Text End-to-End} \\ \cline{2-3}
& None                      & \multicolumn{1}{c}{Full} \\ \hline
\multicolumn{3}{c}{Bounding Box-based methods} \\ \hline      
CharNet \cite{xing2019convolutional}& \multicolumn{1}{c|}{66.6} & \multicolumn{1}{c}{-}   \\ 
ABCNet \cite{liu2020abcnet} & 64.2                      & \multicolumn{1}{c}{75.7}  \\ 
PGNet \cite{wang2021pgnet}  & \multicolumn{1}{c|}{63.1} & \multicolumn{1}{c}{-}   \\ 
Mask TextSpotter \cite{lyu2018mask} & 65.3                      & 77.4                    \\
Qin et al. \cite{qin2019towards}&67.8                       & -  \\
Mask TextSpotter v3 \cite{liao2020masktext}  & 71.2                      & 78.4                  \\ 
MANGO \cite{qiao2021mango}     & \underline{72.9}                      & \textbf{83.6}                  \\ 
PAN++ \cite{wang2021pan++}     & 68.6                      & 78.6                  \\ 
ABCNet v2 \cite{liu2021abcnetv2} & 70.4                      & 78.1                 \\ \hline
\multicolumn{3}{c}{Point-based method} \\ \hline      
\methodName\ (Ours)  & \textbf{74.2} & \underline{82.4}  \\ \hline
\end{tabular}
\end{table}

\begin{table}[t!]
\centering
\caption{End-to-end recognition results on SCUT-CTW1500.  ``None" represents lexicon-free. ``Full" represents that we use all the words appeared in the test set. ABCNet* means using the github checkpoint\protect\footnotemark.}
\label{CTW1500 e2e}
\footnotesize
\begin{tabular}{r|c|c}
\hline
\multirow{2}*{Method} & \multicolumn{2}{c}{SCUT-CTW1500 End-to-End}  \\ \cline{2-3} 
           & None                      & Full  \\ \hline
\multicolumn{3}{c}{Bounding Box-based methods} \\ \hline 
TextDragon \cite{feng2019textdragon} & 39.7                      & 72.4                  \\ 
ABCNet \cite{liu2020abcnet}   & 45.2                      & 74.1 \\
ABCNet* \cite{liu2020abcnet} & 53.2                      & 76.0 \\
MANGO \cite{qiao2021mango}   & \underline{58.9}                      & \underline{78.7}                   \\ 
ABCNet v2 \cite{liu2021abcnetv2}  & 57.5                      & 77.2                   \\ \hline
\multicolumn{3}{c}{Point-based method} \\ \hline 
\methodName\ (Ours) & \textbf{63.6} & \textbf{83.8} \\ \hline
\end{tabular}
\end{table}

\footnotetext{\sf  https://github.com/aim-uofa/AdelaiDet/blob/master/configs/BAText/README.md}

\begin{figure*}[t!]
    \centering
    \begin{subfigure}{0.30\linewidth}
        \centering
        \includegraphics[width=5.2cm, height=2.9cm]{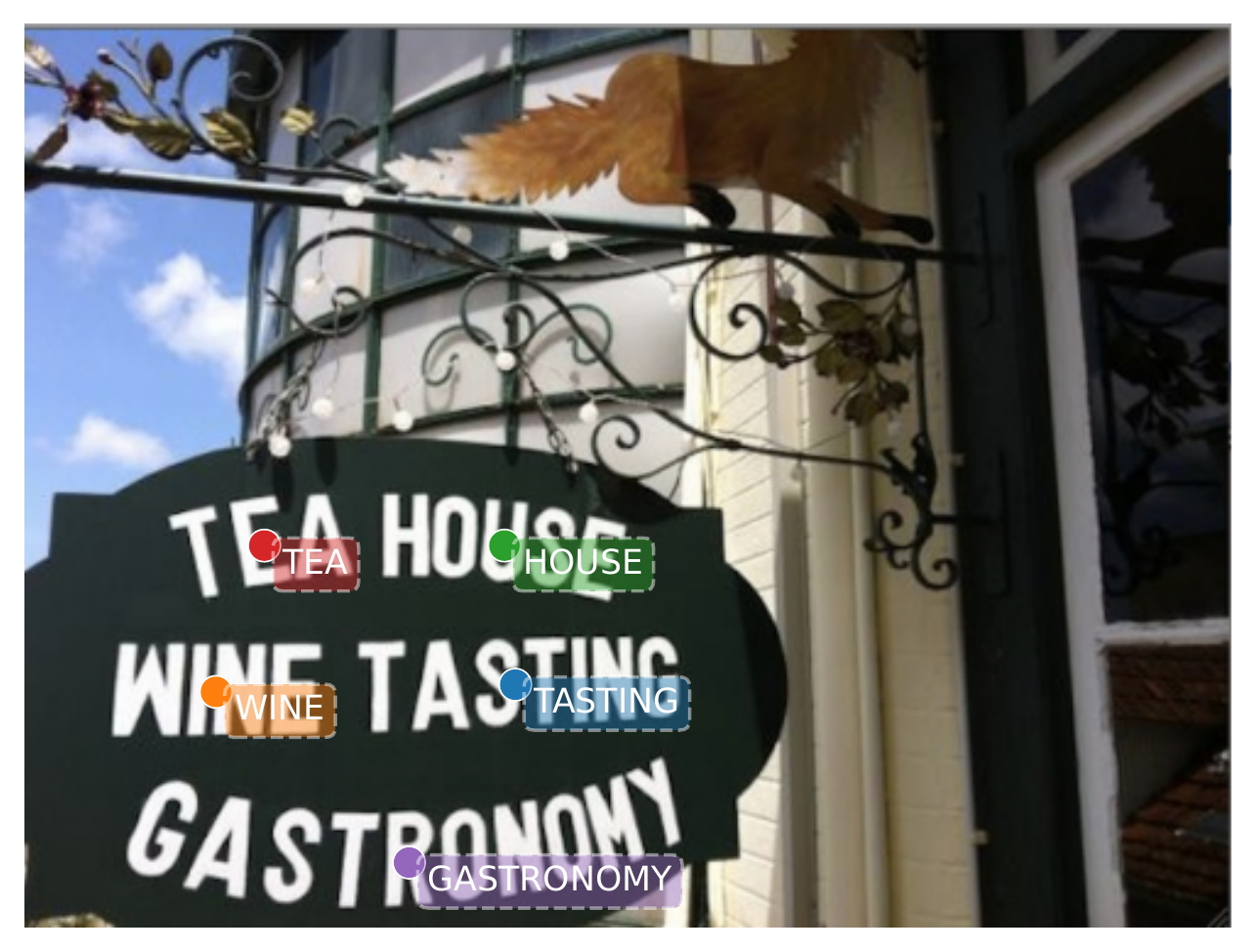}
    \end{subfigure}
    \begin{subfigure}{0.30\linewidth}
        \centering
        \includegraphics[width=5.2cm, height=2.9cm]{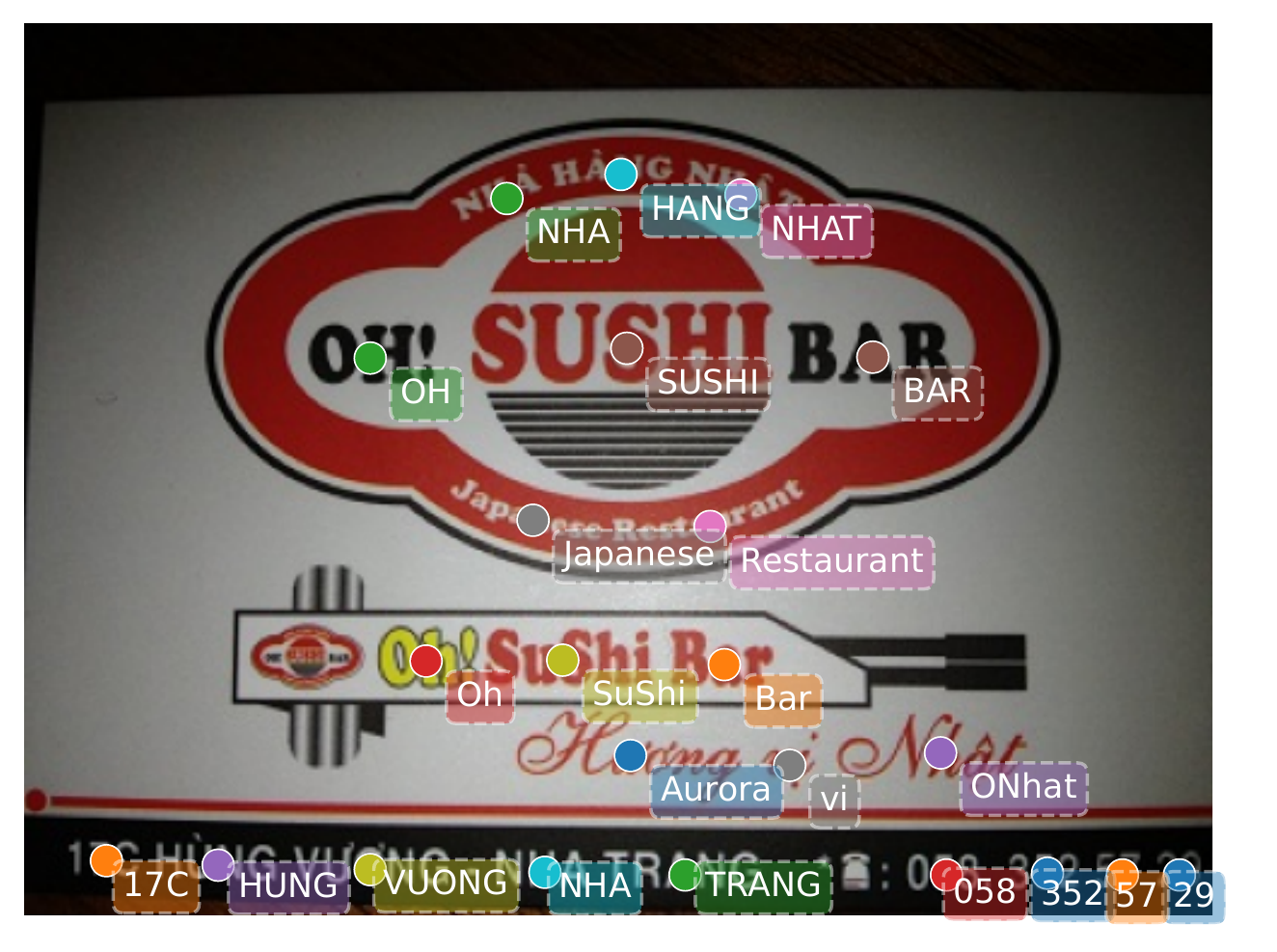}
    \end{subfigure}
    \begin{subfigure}{0.30\linewidth}
        \centering
        \includegraphics[width=5.2cm, height=2.9cm]{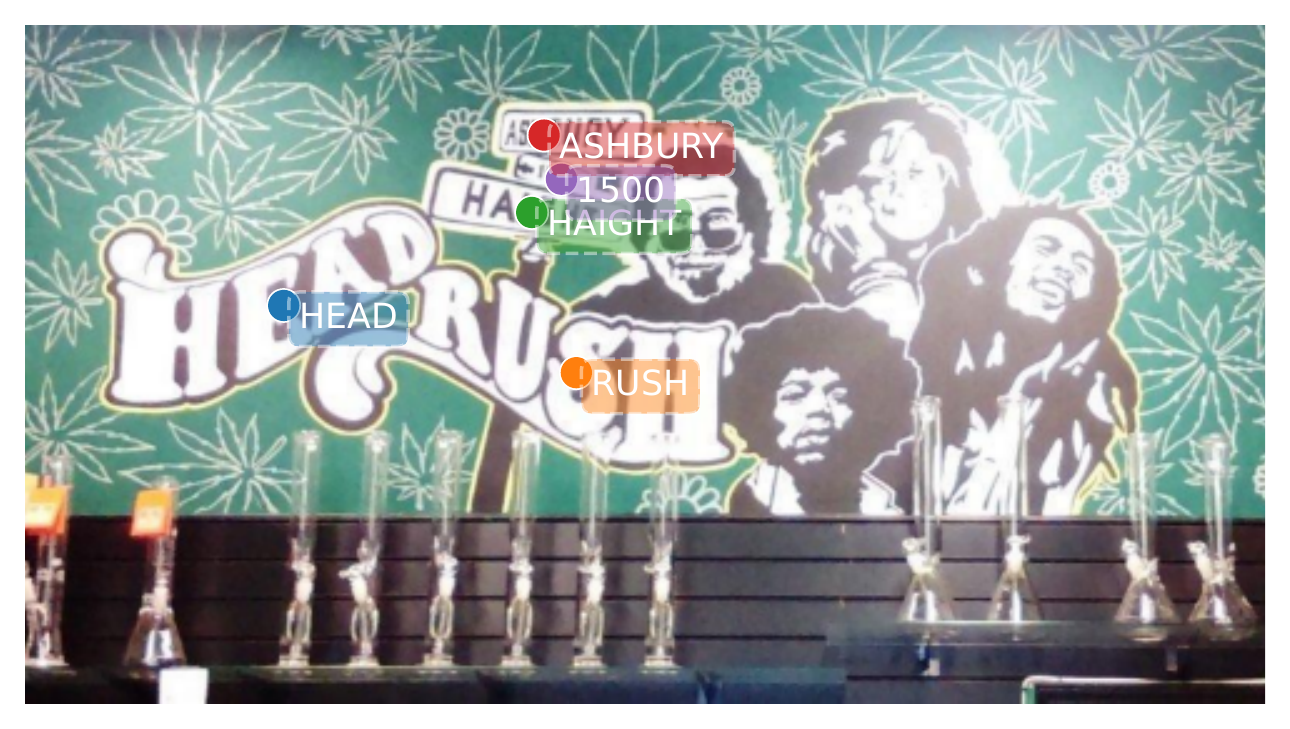}
    \end{subfigure}
    
    \begin{subfigure}{0.30\linewidth}
        \centering
        \includegraphics[width=5.2cm, height=2.9cm]{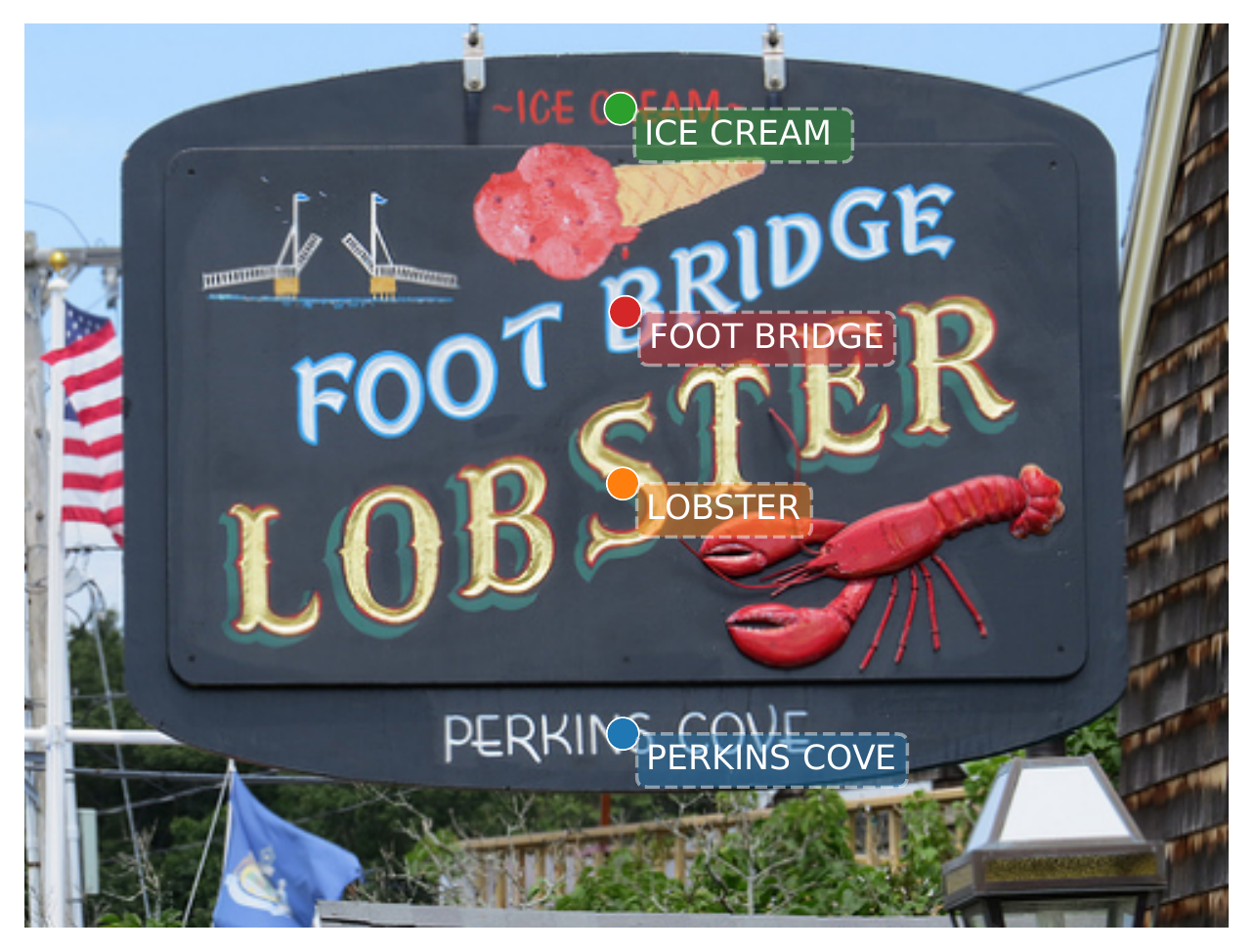}
    \end{subfigure}
    \begin{subfigure}{0.30\linewidth}
        \centering
        \includegraphics[width=5.2cm, height=2.9cm]{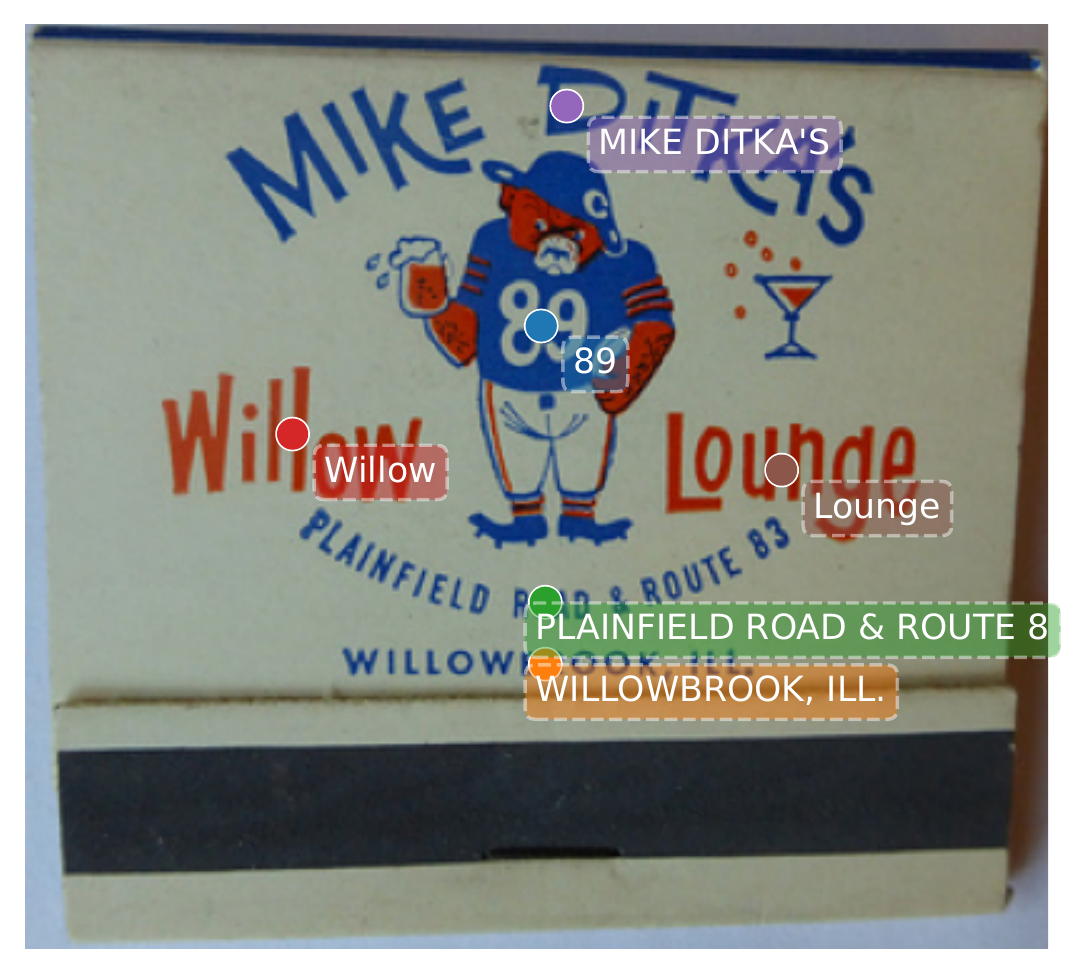}
    \end{subfigure}
    \begin{subfigure}{0.30\linewidth}
        \centering
        \includegraphics[width=5.2cm, height=2.9cm]{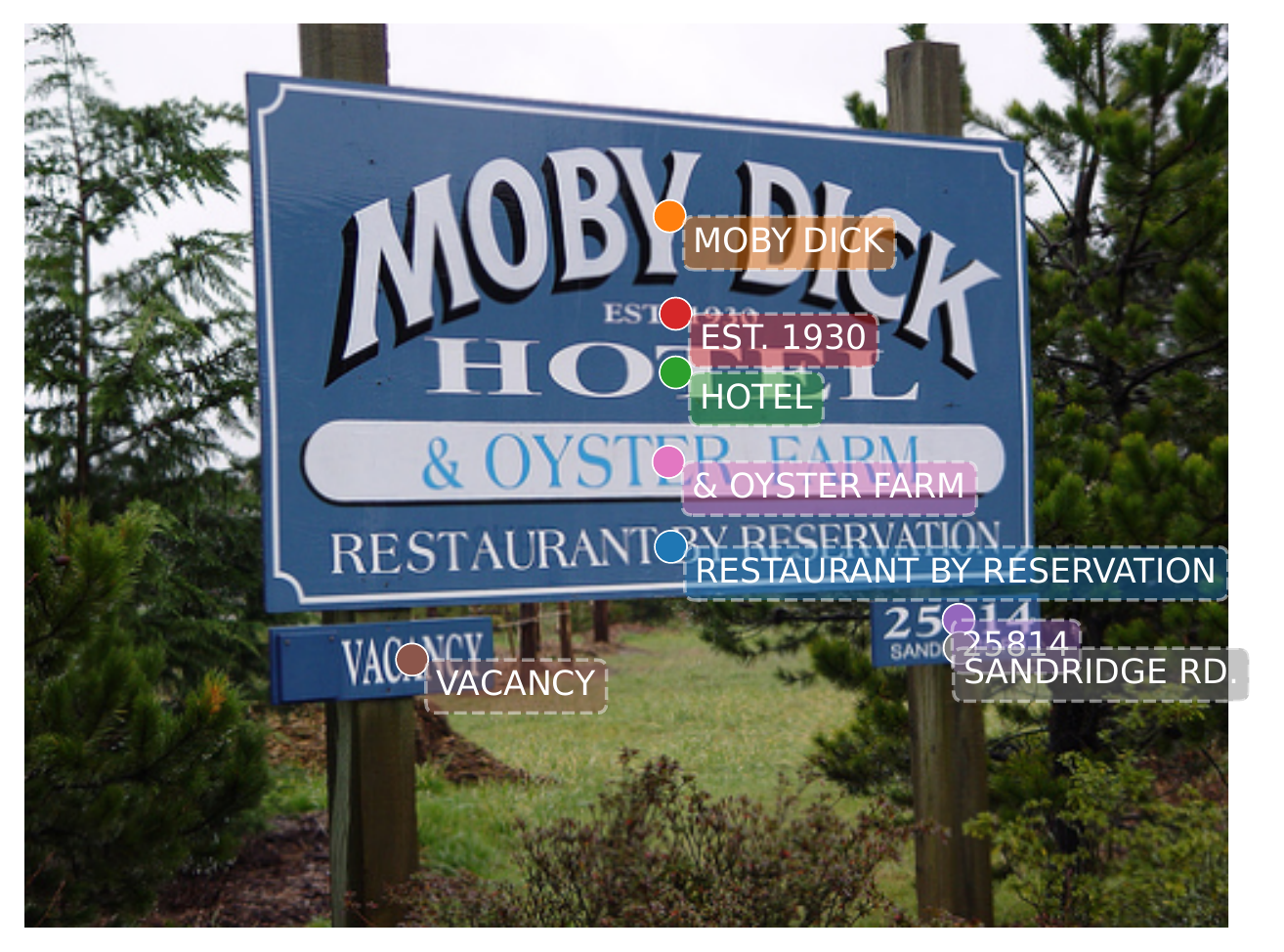}
    \end{subfigure}
    
    \begin{subfigure}{0.30\linewidth}
        \centering
        \includegraphics[width=5.2cm, height=2.9cm]{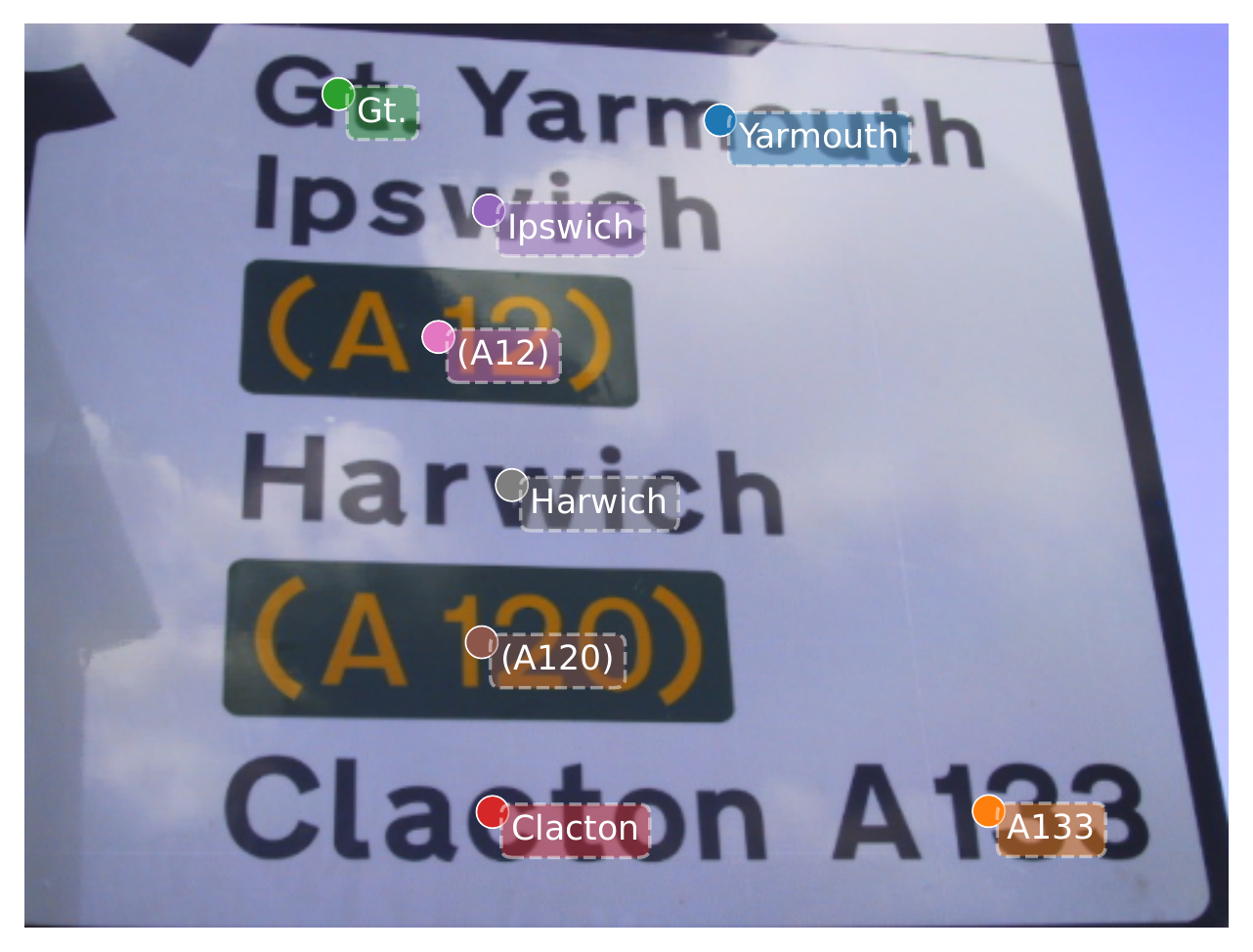}
    \end{subfigure}
    \begin{subfigure}{0.30\linewidth}
        \centering
        \includegraphics[width=5.2cm, height=2.9cm]{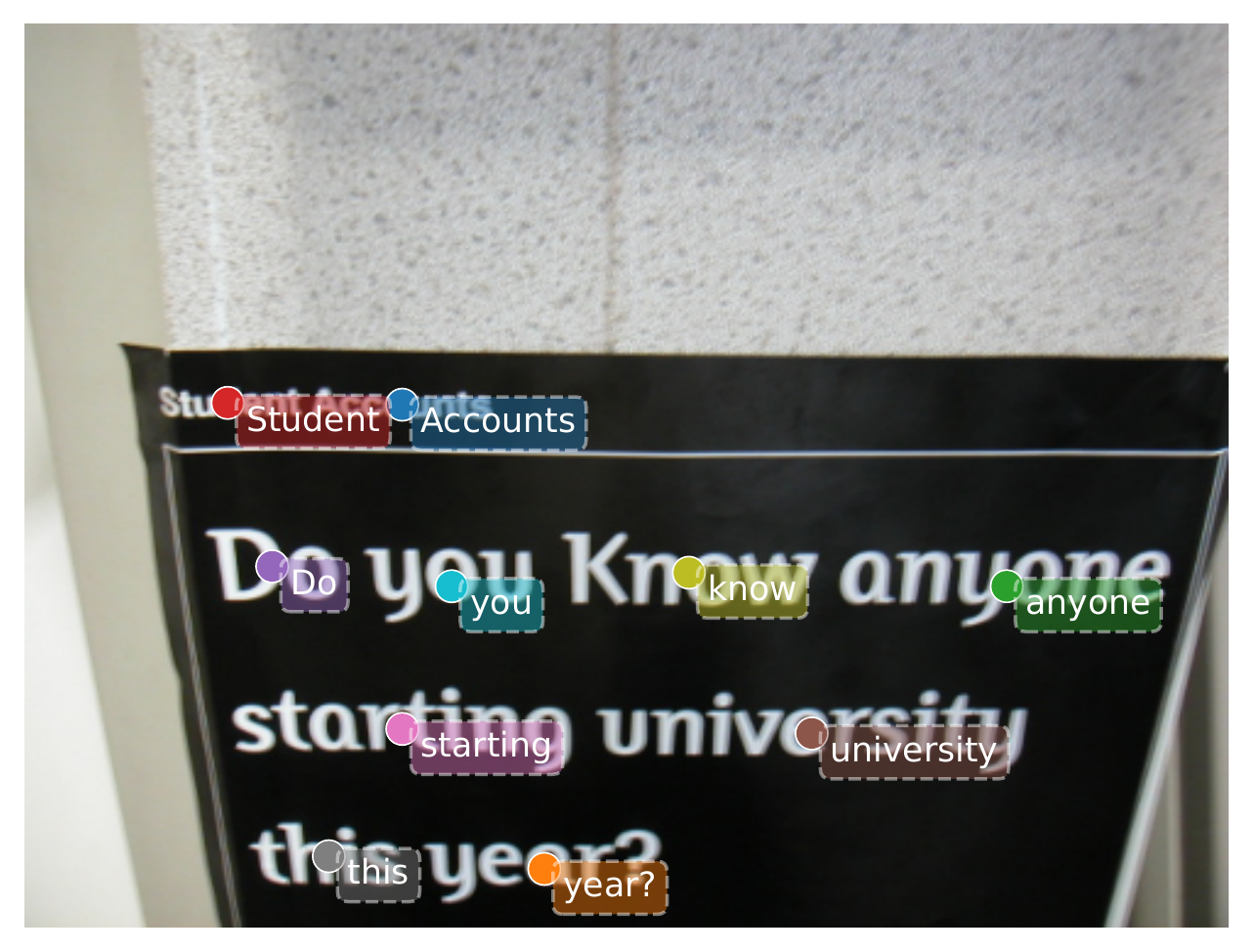}
    \end{subfigure}
    \begin{subfigure}{0.30\linewidth}
        \centering
        \includegraphics[width=5.2cm, height=2.9cm]{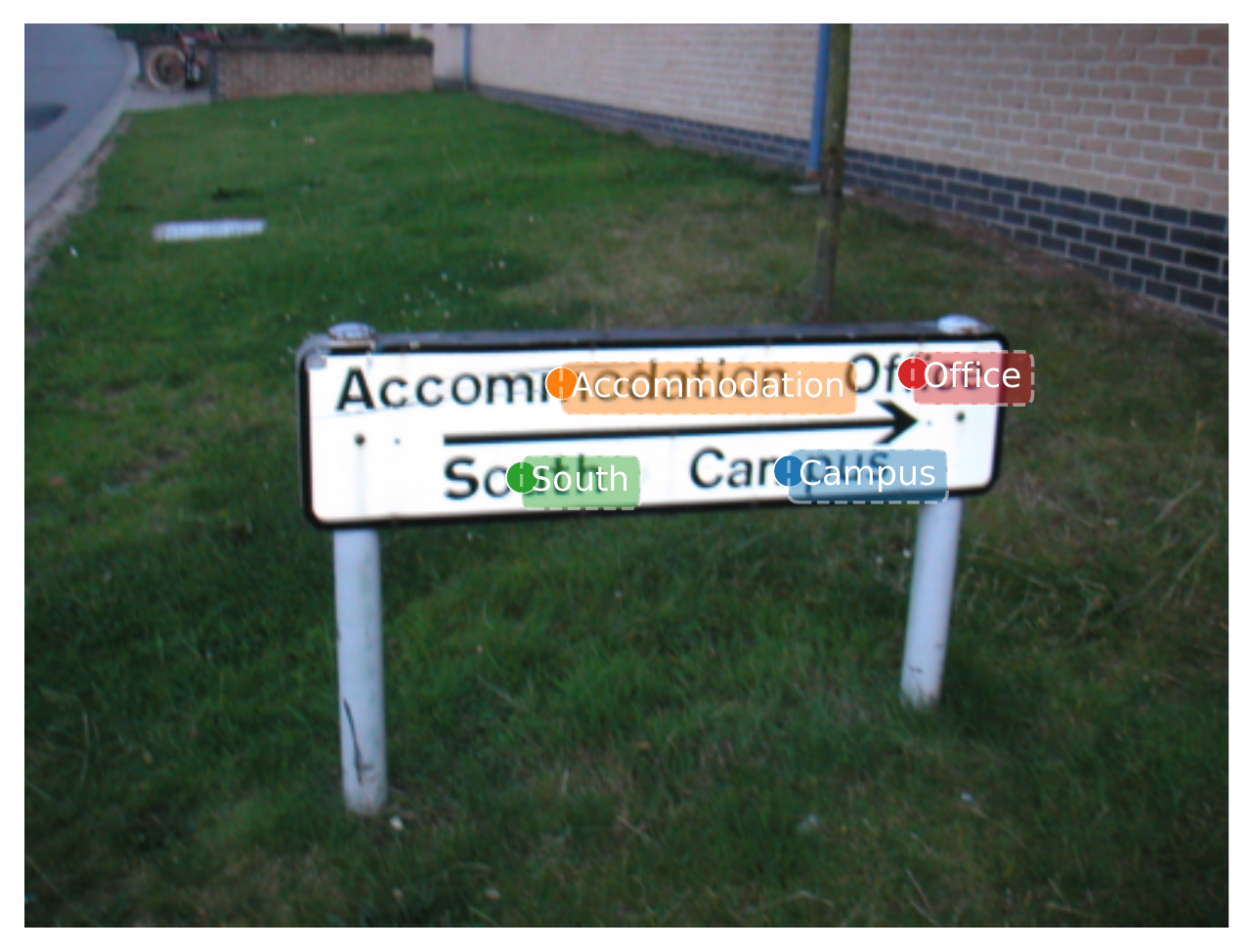}
    \end{subfigure}
    
    \begin{subfigure}{0.30\linewidth}
        \centering
        \includegraphics[width=5.2cm, height=2.9cm]{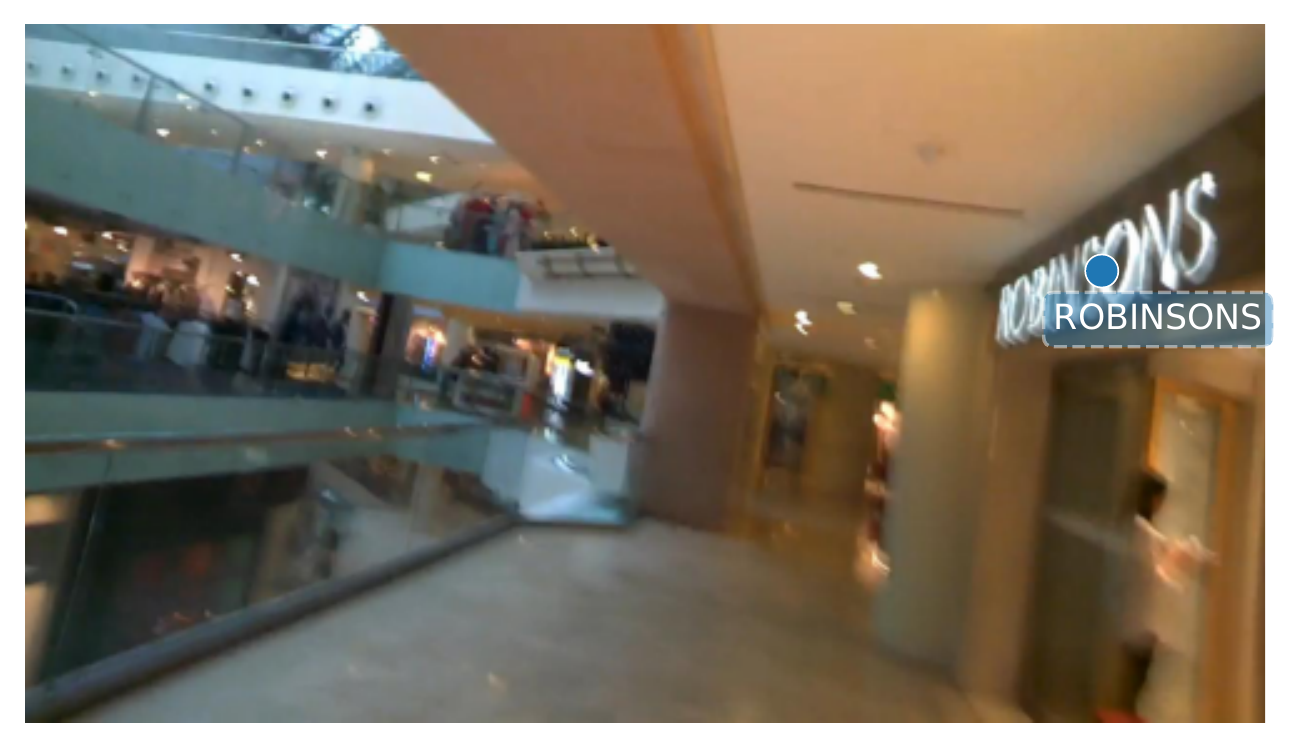}
    \end{subfigure}
    \begin{subfigure}{0.30\linewidth}
        \centering
        \includegraphics[width=5.2cm, height=2.9cm]{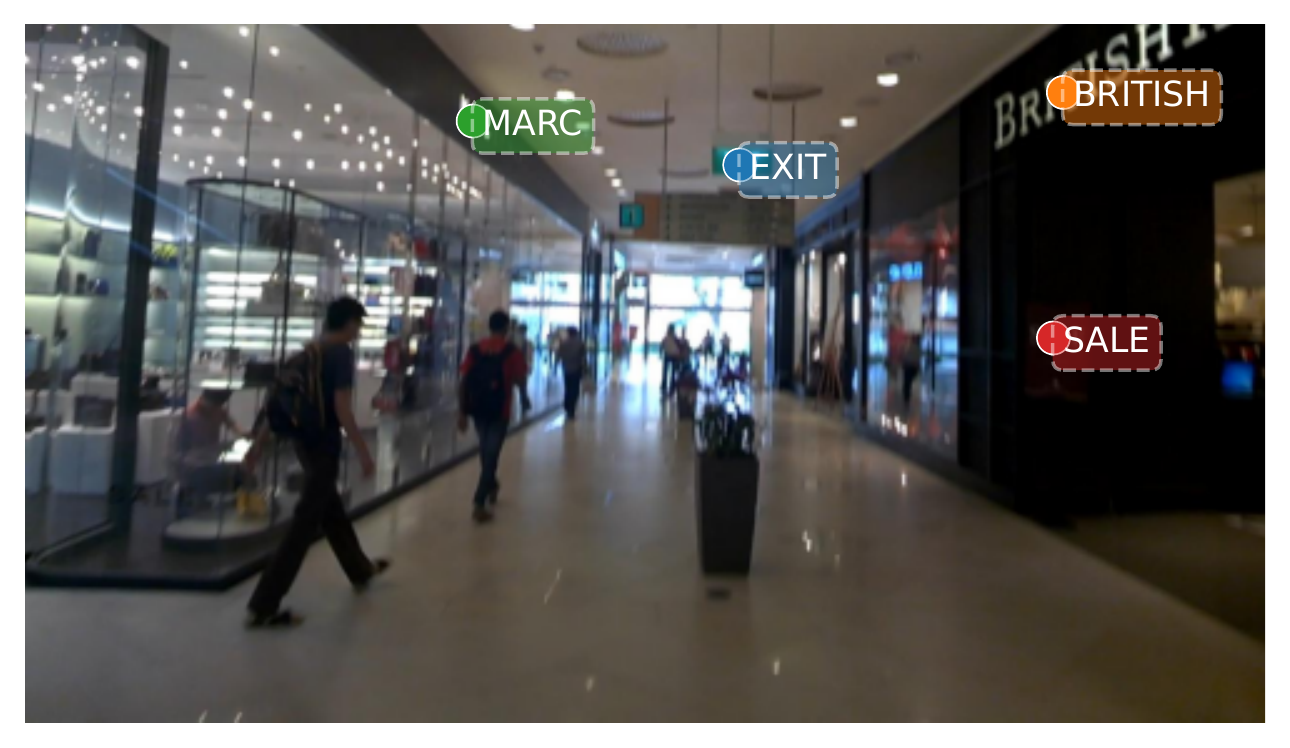}
    \end{subfigure}
    \begin{subfigure}{0.30\linewidth}
        \centering
        \includegraphics[width=5.2cm, height=2.9cm]{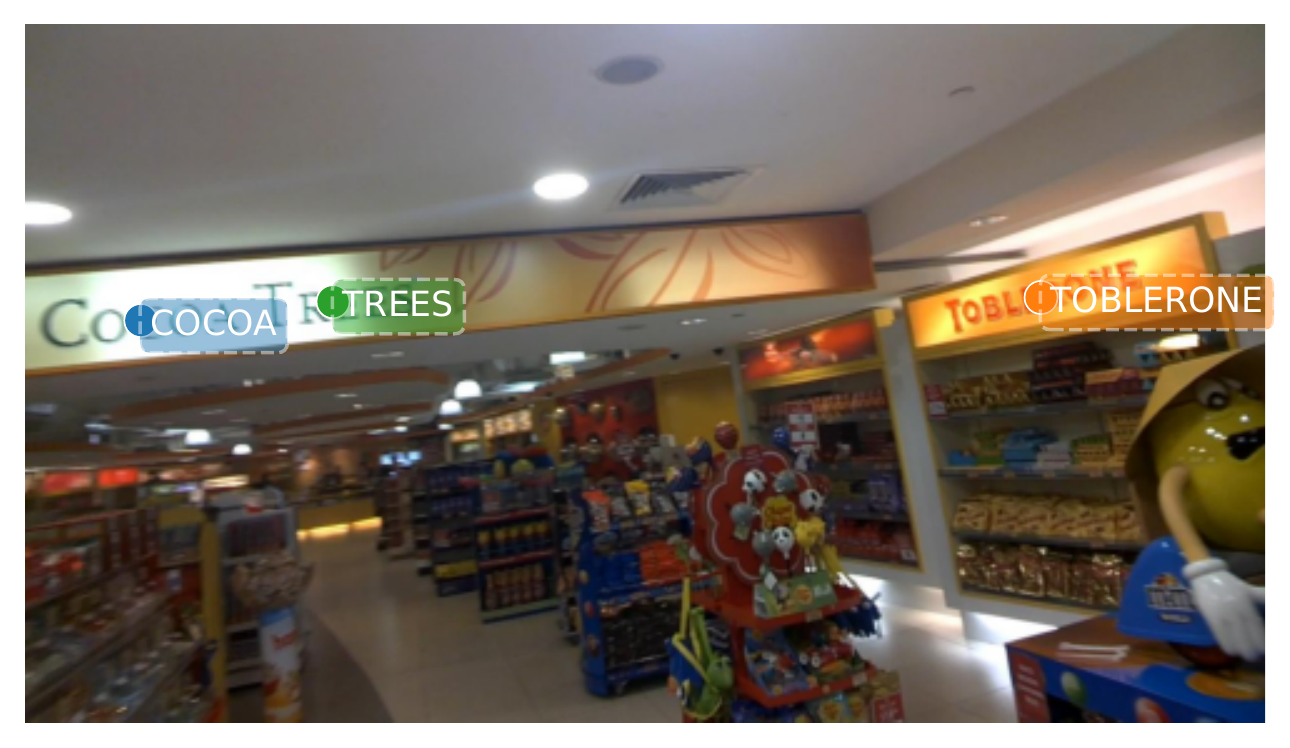}
    \end{subfigure}

    \caption{Qualitative results on the scene text benchmarks. Images are selected from Total-Text (first row), SCUT-CTW1500 (second row), ICDAR 2013 (third row), and ICDAR 2015 (fourth row). Zoom in for best view.}
    \label{fig_vis}
\end{figure*}

\subsubsection{Horizontal-Text Dataset}
Table~\ref{ICDAR 2013 End-to-End recognition result} compares the proposed \methodName\ with existing methods on the widely used ICDAR 2013~\cite{karatzas2013icdar} benchmark. Our method achieves state-of-the-art results with all three lexicons. It should be noted that the proposed \methodName\ only utilizes a single-point for training, while the other approaches are fully trained with more costly bounding boxes.

\subsubsection{Multi-Oriented Dataset}\label{sec:multi-orientd dataset}
The quantitative results of the ICDAR 2015~\cite{karatzas2015icdar} dataset are shown in Table~\ref{ICDAR 2015 End-to-End recognition result}. A performance gap between the proposed \methodName\ and state-of-the-art methods can still be found, which shows some limitations of our method for tiny texts that are often presented in the ICDAR 2015 dataset. Because the sequence is directly decoded from the feature of the entire image without dedicated RoI operations, the tiny texts are difficult to handle with our method.
        
\subsubsection{Arbitrarily Shaped Dataset}
We further compare our method with existing approaches on the benchmarks containing arbitrarily shaped texts, including Total-Text~\cite{ch2017total} and SCUT-CTW1500~\cite{liu2019curved}. As shown in Table~\ref{Total-Text e2e}, \methodName\ achieves state-of-the-art performance only using extremely low-cost point annotations. Additionally, Table~\ref{CTW1500 e2e} shows that our method outperforms state-of-the-art approaches by a large margin on the challenging SCUT-CTW1500 dataset, which further identifies the potentiality of our method.

\begin{table*}[t!]
    \centering
    \caption{Comparison between the end-to-end recognition results of the SPTS and NPTS models.}
    \label{SPTS-NoPoint}
    \small
    \begin{tabular}{r|cc|cc|ccc|ccc}
        \hline
        \multirow{2}*{Method} & \multicolumn{2}{c|}{Total-Text} & \multicolumn{2}{c|}{SCUT-CTW1500} & \multicolumn{3}{c|}{ICDAR 2013} & \multicolumn{3}{c}{ICDAR 2015} \\
        \cline{2-11}
         & None & Full & None & Full & S & W & G & S & W & G \\
        \hline
        SPTS & \textbf{74.2} & \textbf{82.4} & \textbf{63.6} & \textbf{83.8} & \textbf{93.3} & \textbf{91.7} & \textbf{88.5} & \textbf{77.5} & \textbf{70.2} & \textbf{65.8} \\
        NPTS & 64.7 & 71.9 & 55.4 & 74.3 & 90.4 & 84.9 & 80.2 & 69.4 & 60.3 & 55.6 \\
        \hline
    \end{tabular}
\end{table*}

\subsubsection{Summary}        
In summary, the proposed \methodName\ can achieve state-of-the-art performance compared with previous text spotters on several widely used benchmarks. Especially on the two curved datasets, \emph{i.e.}, Total-Text~\cite{ch2017total} and SCUT-CTW1500~\cite{liu2019curved}, the proposed \methodName\ outperforms some recently proposed methods by a large margin. The reason why our methods can achieve better accuracy on arbitrary-shaped texts might be: (1) The proposed \methodName\ discards the task-specific modules (\emph{e.g.}, RoI modules) designed based on prior knowledge; therefore, the recognition accuracy is decoupled with the detection results, \emph{i.e.}, \methodName\ can achieve acceptable recognition results even the detection position is shifted. 
However, the recognition heads of other methods heavily rely on the detection results, which is the main reason of their poor end-to-end accuracy. Once the text instance cannot be perfectly localized, their recognition heads fail to work. 
(2) Although previous models are trained in an end-to-end manner, the interactions between their detection and recognition branches are limited. Specifically, the features fed to the recognition module are sampled based on the ground-truth position while training but from detection results at the inference stage, leading to feature misalignment, which is far more severe on curved text instances. However, by tackling the spotting task in a sequence modeling manner, the proposed \methodName\ eliminates such issues, thus showing more robustness on arbitrarily shaped datasets. The visualization results of SPTS on four testing datasets are shown in Fig. \ref{fig_vis}.

\begin{figure}[t!]
    \centering
    \includegraphics[width=0.9\linewidth]{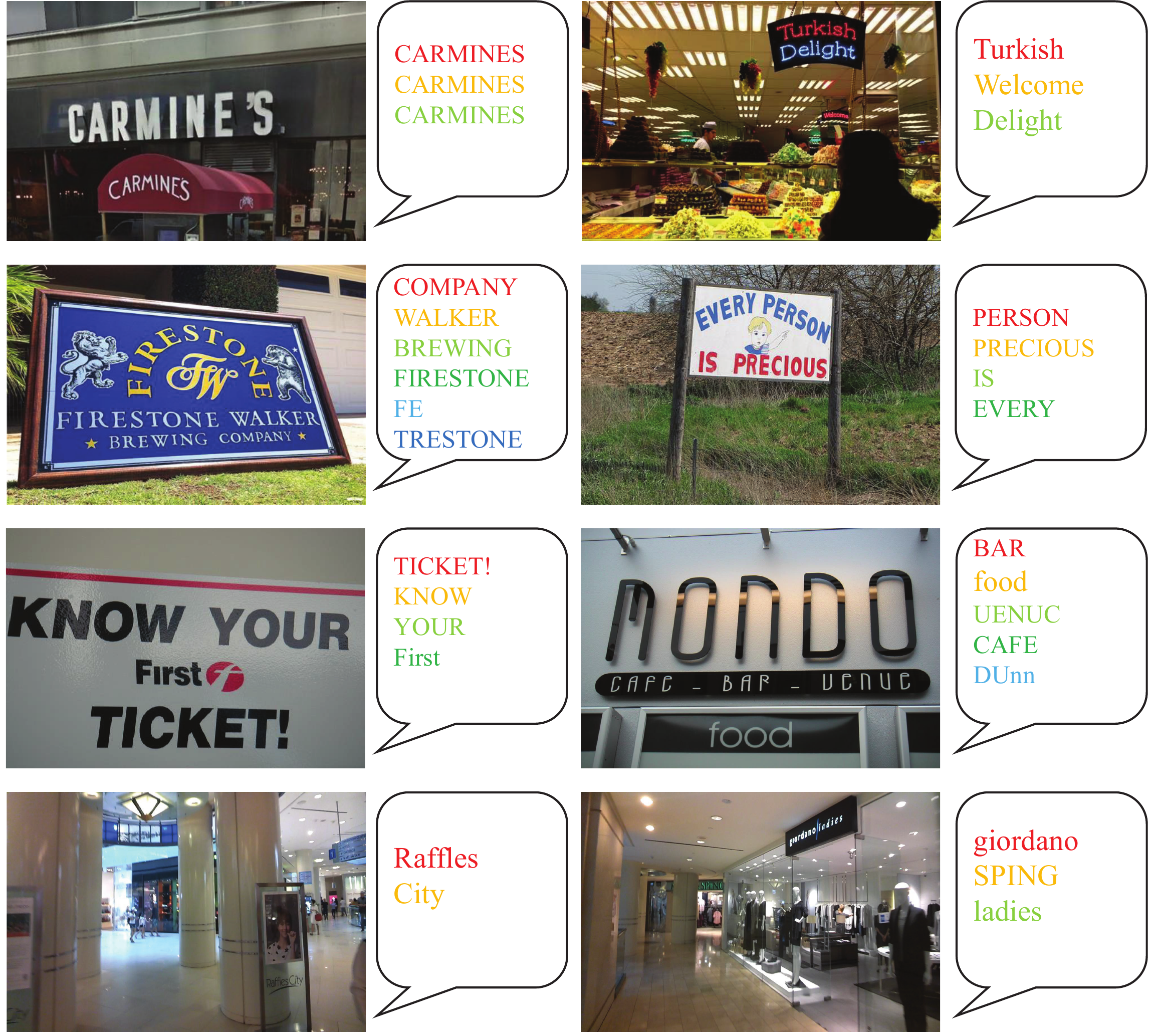}
    \caption{Qualitative results of the NPTS model on several scene text benchmarks. Images are selected from Total-Text (first row), SCUT-CTW1500 (second row), ICDAR 2013 (third row), and ICDAR 2015 (fourth row). Zoom in for best view.}
    \label{fig:vis_wopoint}
\end{figure}

\subsection{Extensions of SPTS}
\subsubsection{No-Point Text Spotting}
The experiments suggest that the detection and recognition may have been decoupled. Based on the results, we further show that SPTS can be converged even without the supervision of the single point annotations. The No-Point Text Spotting (NPTS) model is obtained by removing the coordinates of the indicated points from the constructed sequence. Fig. \ref{fig:vis_wopoint} shows the qualitative results of NPTS, which indicates the model may have learned the ability to implicitly find out the locations of the text merely based on the transcriptions. The comparison between the end-to-end recognition results of the SPTS and NPTS models is presented in Tab. \ref{SPTS-NoPoint}. 
The evaluation metric described in Sec. \ref{subsec:eval_protocol} is adapted for NPTS, where the distance matrix between the predicted and GT points is replaced with an edit distance matrix between the predicted and GT transcriptions. 
Despite the obvious gap between SPTS and NPTS, the preliminary results achieved by NPTS are still surprising and very encouraging, which is worth studying in the future.

\subsubsection{Single-Point Object Detection}
\begin{figure}[t!]
\centering
    \begin{subfigure}{0.4\linewidth}
        \includegraphics[width=3.2cm, height=2.4cm]{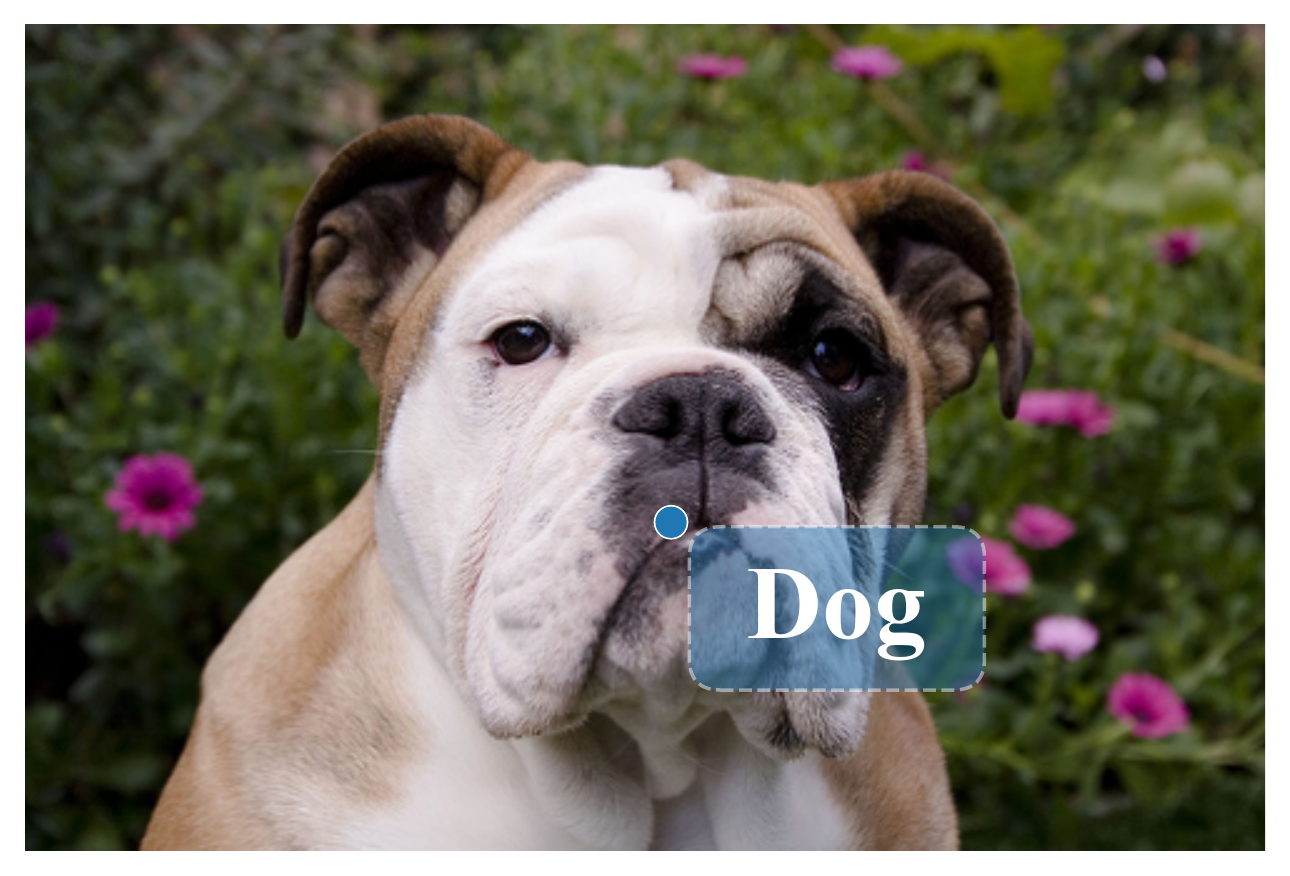}
    \end{subfigure}
    \begin{subfigure}{0.4\linewidth}
        \includegraphics[width=3.2cm, height=2.4cm]{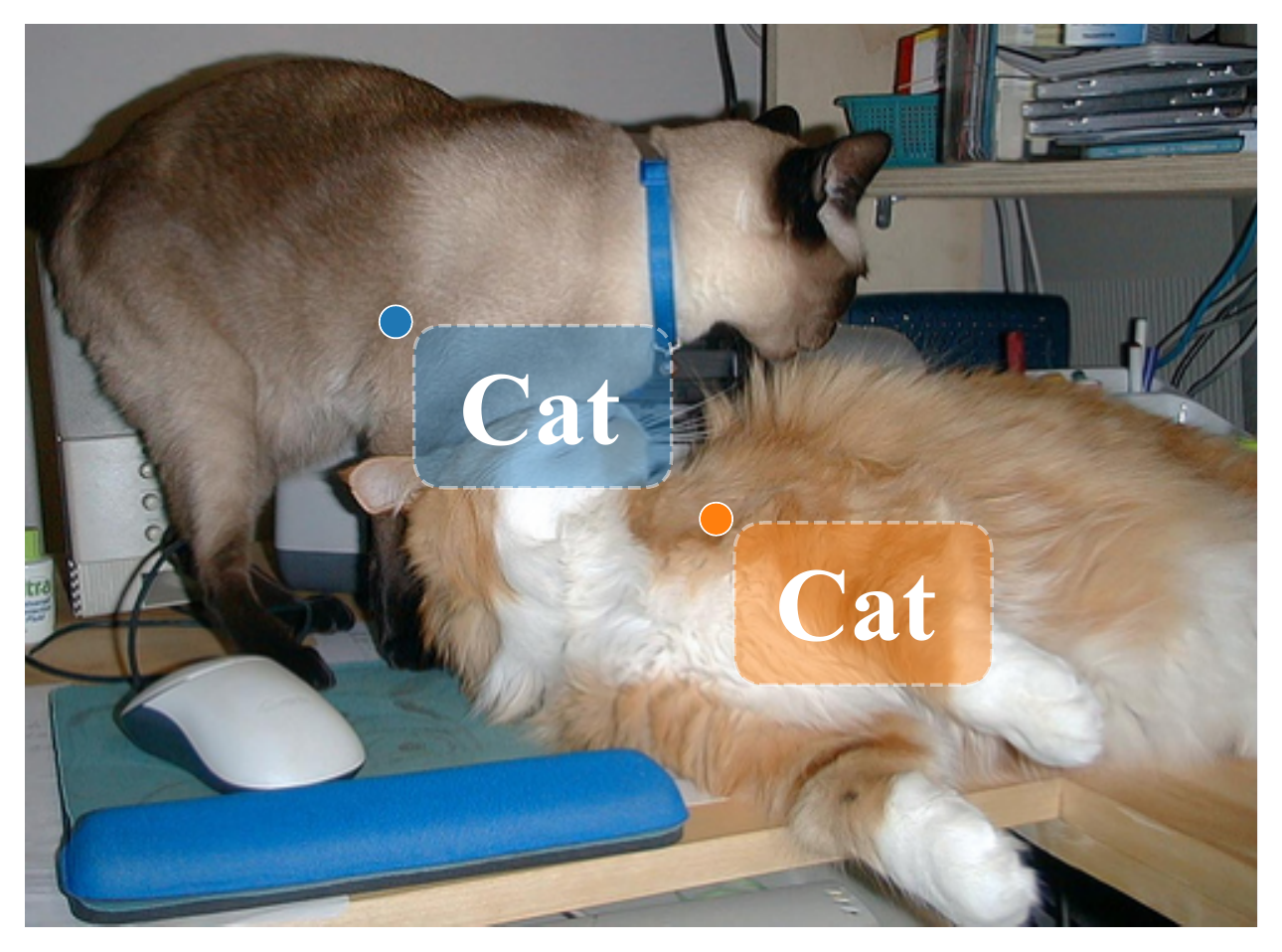}
    \end{subfigure}
    \caption{Qualitative object detection results on the Pascal VOC 2012 validation set under the single-point supervision. }
    \label{fig:object_detection}
\end{figure}

To demonstrate the generality of SPTS, we conduct experiments on the Pascal VOC~\cite{everingham2010pascal} object detection task, where the model is trained with central points and the corresponding categories. All other settings are identical to the text spotting experiment. Some preliminary qualitative results on the validation set are shown in Fig.~\ref{fig:object_detection}. The results suggest that single-point might be viable to provide extremely low-cost annotation for general object detection.


\section{Limitation}
\label{sec:discussion}

One limitation of the proposed framework is that the training procedure requires a large number of computing resources. For example, 150 epochs for 160k scene text pretraining and 200 epochs for fine-tuning require approximately 63 hours when the model is distributively trained using 32 NVIDIA V100 GPU cards. Moreover, owing to the auto-regressive decoding manner, the inference speed is only 1.0 fps on ICDAR 2013 using one NVIDIA V100 GPU.

Another limitation of SPTS is that it can not well handle extremely tiny text spotting (as indicated in Sec. \ref{sec:multi-orientd dataset}), because the feature representation ability of SPTS can not reach a high resolution for extracting effective tiny text features. This issue deserves further study in the future.

\section{Conclusion}
We propose SPTS, which is, to the best of our knowledge, a pioneering method that tackles scene text spotting using only the extremely low-cost single-point annotation. The successful attempt sheds some brand-new insights that challenges the necessity of the traditional box annotations in the field.
SPTS is an auto-regressive Transformer-based framework that can simply generate the results as sequential tokens, which can avoid complex post-processing or exclusive sampling stages. 
Based on such a concise framework, extensive experiments demonstrate state-of-the-art performance of SPTS on various datasets.
We further show that our method has the potential to be extended to the no-point text spotting and generic object detection tasks. 

\section*{Acknowledgement}
This research is supported in part by NSFC (Grant No.: 61936003), GD-NSF (no.2017A030312006, No.2021A1515011870), and the Science and Technology Foundation of Guangzhou Huangpu Development District (Grant 2020GH17).

\bibliographystyle{ACM-Reference-Format}
\balance
\bibliography{egbib.bib}

\clearpage
\captionsetup{font={normal}}
\balance
\appendix
\noindent\textbf{\Huge Appendix}
\section{Tiny Text Spotting}
As discussed in Sec. \ref{sec:multi-orientd dataset} and Sec. \ref{sec:discussion}, the proposed SPTS can not accurately recognize tiny texts because it directly predicts the sequence based on the low-resolution high-level features without RoI operations. Especially on the ICDAR 2015 dataset, there still is a performance gap (65.8 vs. 74.2)  between our method and the state-of-the-art approaches. Quantitatively, if the texts with area smaller than 3000 (after resizing) are ignored during evaluation, the F-measure with generic lexicons on ICDAR 2015 will be significantly improved from 65.8 to 73.5. Furthermore, current state-of-the-art methods on ICDAR 2015 usually adopt larger image size during testing. For example, the short sides of the testing images are resized to 1440 pixels while the long sides are shorter than 4000 pixels in Mask TextSpotterV3 \cite{liao2020masktext}. As shown in Tab. \ref{ICDAR 2015 End-to-End recognition result v2}, the performance of SPTS on ICDAR 2015 with larger testing size is much better than that with smaller testing size, indicating the tiny text spotting is an important issue worth studying in the future.

\begin{table}[b!]
    \centering
    \caption{End-to-end recognition results on ICDAR 2015. “S”, “W”, and “G” represent recognition with “Strong”, “Weak”, and “Generic” lexicon, respectively. Bold indicates the state of the art and underline indicates the second best.}
    \label{ICDAR 2015 End-to-End recognition result v2}
    \small
    \begin{tabular}{r|c|c|c}
    \hline
    \multirow{2}{*}{Method} & \multicolumn{3}{c}{IC15 End-to-End}                    \\ \cline{2-4} 
                            & \multicolumn{1}{c|}{S}    & \multicolumn{1}{c|}{W}    & G    \\ \hline
    \multicolumn{4}{c}{Bounding Box-based methods} \\ \hline                               
    FOTS \cite{liu2018fots}                    & \multicolumn{1}{c|}{81.1} & \multicolumn{1}{c|}{75.9} & 60.8 \\
    Mask TextSpotter \cite{liao2019mask}        & \multicolumn{1}{c|}{83.0} & \multicolumn{1}{c|}{77.7} & \underline{73.5} \\
    CharNet \cite{xing2019convolutional}                 & \underline{83.1} & \multicolumn{1}{c|}{\textbf{79.2}} & 69.1 \\
    TextDragon \cite{feng2019textdragon} & \multicolumn{1}{c|}{82.5} & \multicolumn{1}{c|}{78.3} & 65.2 \\
    Mask TextSpotter v3 \cite{liao2020masktext} & \multicolumn{1}{c|}{\textbf{83.3}} & \multicolumn{1}{c|}{78.1} & \textbf{74.2} \\
    MANGO \cite{qiao2021mango}                   & \multicolumn{1}{c|}{81.8} & \underline{78.9} & 67.3 \\
    ABCNetV2 \cite{liu2021abcnetv2}                & \multicolumn{1}{c|}{82.7} & \multicolumn{1}{c|}{78.5} & 73.0 \\ 
    PAN++ \cite{wang2021pan++}                   & \multicolumn{1}{c|}{82.7} & \multicolumn{1}{c|}{78.2} & 69.2 \\
    \hline
    \multicolumn{4}{c}{Point-based method} \\ \hline   
    \methodName\ (1000) &  77.5 & 70.2 & 65.8 \\
    \methodName\ (1440) &  79.5 & 74.1 & 70.2 \\ \hline
    \end{tabular}
\end{table}


\section{Order of Text Instances}

As described in Sec. \ref{sec:seq construction}, the text instances are randomly ordered in the constructed sequence following Pix2Seq \cite{chen2021pix2seq}. In this section, we further investigate the impact of the order of text instances. The performances on Total-Text and SCUT-CTW1500 of different ordering strategies are presented in Tab. \ref{tab_ordering}. The ``Area'' and ``Dist2ori'' mean that text instances are sorted by the area and the distance to the topleft origin in descending order, respectively. The ``Topdown'' indicates that text instances are arranged from top to bottom. It can be seen that the random order adopted in SPTS achieves the best performance, which may be explained by the improved robustness due to the different sequences constructed for the same image at different iterations.

\begin{table}[b!]
    \centering
    \caption{Ablation study of different ordering strategies of text instances in the sequence construction.}
    \label{tab_ordering}
    \small
    \begin{tabular}{r|cc|cc}
        \hline
        \multirow{2}{*}{Order} & \multicolumn{2}{c|}{Total-Text} & \multicolumn{2}{c}{SCUT-CTW1500} \\
        \cline{2-5}
         & None & Full & None & Full \\
        \hline
        Area & 70.7 & 79.2 & 59.0 & 75.3 \\
        Topdown & 73.2 & 81.3 & 62.7 & 83.7 \\
        Dist2ori & 72.1 & 81.8 & 61.1 & 79.6 \\
        \hline 
        Random & \textbf{74.2} & \textbf{82.4} & \textbf{63.6} & \textbf{83.8}  \\
        \hline
    \end{tabular}
\end{table}

\section{Further Discussion on the Representation of Text Instances}

In Sec. \ref{sec_rect_bezier}, the ablation experiments demonstrate that the SPTS-Point with fewer parameters for describing the location of text instances outperforms SPST-Rect and SPTS-Bezier. The longer sequence required by SPTS-Rect and SPTS-Bezier may make them difficult to converge. In Tab. \ref{tab:ab_box_shape}, the results of SPTS-Rect and SPTS-Bezier are obtained using the same training schedule as SPTS-Point. To further explore their potentiality, we compare the SPTS-Bezier 
trained for $2\times$ epochs with SPTS-Point in Tab. \ref{tab_2x_iter}. Normally, the SPTS-Bezier should perform better than SPTS-Point owing to the more detailed annotation. However, it can be seen that the SPTS-Bezier with $2\times$ epochs does not significantly outperform the counterpart with $1\times$ epochs and is still inferior to the SPTS-Point with $1\times$ epochs. The reason may be the dramatically increased difficulty of the convergence of the Transformer with longer decoded sequences. More training data and epochs may be necessary to address this problem.

\begin{table}[b!]
    \centering 
    \caption{Further comparison of different representations of text instances.}
    \label{tab_2x_iter} 
    \small 
    \begin{tabular}{r|c|cc|cc|c}
        \hline
        \multirow{2}*{Variants} & \multirow{2}*{Epochs} & \multicolumn{2}{c|}{Total-Text} & \multicolumn{2}{c|}{SCUT-CTW1500} & \multirow{2}*{$N_p$} \\
        \cline{3-6}
        & & None & Full & None & Full & \\
        \hline 
        SPTS-Bezier & $1\times$ & 60.6 & 71.6 & 52.6 & 73.9 & 16 \\
        SPTS-Bezier & $2\times$ & 62.9 & 74.4 & 51.1 & 74.3 & 16 \\
        \hline 
        SPTS-Point  & $1\times$ & \textbf{74.2} & \textbf{82.4} & \textbf{63.6} & \textbf{83.8} & 2 \\
        \hline
    \end{tabular}
\end{table}








\end{document}